%% file: main.tex
\documentclass[runningheads]{llncs}

\usepackage{eccv}

\input{math_commands.tex}

\usepackage{eccvabbrv}

\usepackage{graphicx}
\usepackage{booktabs}

\usepackage{tikz}
\usepackage{comment}
\usepackage{amsmath,amssymb} %
\usepackage{color}

\usepackage{epsfig}
\usepackage{url}
\usepackage[utf8]{inputenc} %
\usepackage[T1]{fontenc}    %
\usepackage{subcaption}
\usepackage[normalem]{ulem} 
\usepackage[export]{adjustbox}
\usepackage{amsfonts}       %
\usepackage{nicefrac}       %
\usepackage{microtype}      %
\usepackage{gensymb}
\usepackage{tabularx}
\usepackage{siunitx}
\usepackage{enumitem}
\usepackage{lineno}
\usepackage{pifont}
\usepackage{mathtools}
\usepackage{bbm}
\usepackage{wrapfig}

\usepackage[font=small,labelfont=bf]{caption}

\usepackage[accsupp]{axessibility}  %

\usepackage{hyperref}

\usepackage{orcidlink}

\definecolor{navy}{rgb}{0.7, 0.1, 0.7}
\definecolor{burgundy}{RGB}{144,0,32}

\newcommand{\KGnote}[1]{{\color{magenta}{\bf KG: }#1}} %
\newcommand{\ZAnote}[1]{{\color{teal}{\bf ZA: }#1}}

\newcommand{\SM}[1]{{\color{brown}#1}} %
\newcommand{\SMnote}[1]{{\color{burgundy}{\bf SM: }#1}}

\renewcommand{\KGnote}[1]{{\color{magenta}}} %
\renewcommand{\ZAnote}[1]{{\color{teal}}}

\renewcommand{\SM}[1]{{\color{black}#1}} %
\renewcommand{\SMnote}[1]{{\color{burgundy}}}

\newcommand{\mc}{\mathcal}

\begin{document}

\title{MistExit: Learning to Exit for Early\\ Mistake Detection in \SM{Procedural} Videos
}

\titlerunning{MistExit}

\author{Sagnik Majumder$^{1}$ \hspace{3mm} Anish Nethi$^{1}$ \hspace{3mm} Ziad Al-Halah$^{2}$ \hspace{3mm} Kristen Grauman$^{1}$}

\authorrunning{Majumder et al.}

\institute{$^1$UT Austin \hspace{3mm} $^2$University of Utah }

\maketitle

\input{sections/abstract}
\input{sections/introduction}

\input{sections/related_work}

\input{sections/task}
\input{sections/approach}
\input{sections/experiments}
\input{sections/conclusion}

\newpage
\bibliographystyle{splncs04}
\bibliography{main}

\newpage
\input{sections/supp}

\end{document}

%% file: math_commands.tex
\usepackage{amsmath,amsfonts,bm}

\def\eqref#1{equation~\ref{#1}}

\def\1{\bm{1}}

\DeclareMathAlphabet{\mathsfit}{\encodingdefault}{\sfdefault}{m}{sl}
\SetMathAlphabet{\mathsfit}{bold}{\encodingdefault}{\sfdefault}{bx}{n}

%% file: sections/abstract.tex
\begin{abstract}
We introduce the task of early mistake detection in video, where the goal is to determine whether a keystep in a procedural activity is performed correctly %
while observing as little of the streaming video as possible. To tackle this problem, we propose a method comprising a mistake detector and a reinforcement learning policy. At each timestep, the detector processes recently observed frames to estimate the keystep’s correctness while anticipating future visual features, enabling reliable early mistake estimates. Meanwhile, the policy aggregates the detector outputs and visual observations over time and adaptively decides when to exit (\ie, stop processing incoming frames) while producing the final prediction. 
Using diverse real-world procedural video datasets, we demonstrate that our MistExit model achieves superior mistake detection accuracy while reducing the fraction of video observed compared to %
state-of-the-art models. Project: \url{https://vision.cs.utexas.edu/projects/mist_exit}.
\end{abstract}

%% file: sections/introduction.tex
\section{Introduction}\label{sec:intro}
\begin{figure*}[t] 
    \centering
    \includegraphics[width=1\linewidth]{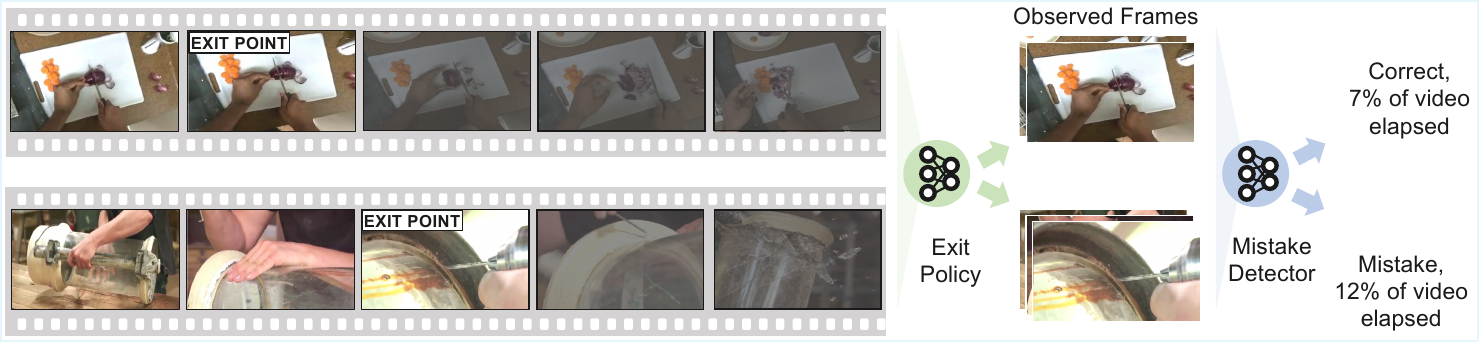} 
    \vspace*{-0.2in}
    \caption{
   Our goal is to learn a policy that, given a streaming video of a keystep in a procedural activity, decides how much of the video to observe before exiting (\ie, stopping inference), such that a mistake detector conditioned on the recently observed frames can accurately determine whether the keystep is a mistake or not while minimizing the fraction of the video observed.  
   \Eg, in a video like the one shown in row 2, a well-trained model may observe a drill bit (frame 3) and infer that the step will eventually result in a mistake---the glass breakage in later frames confirms this---and exit promptly, thereby using only about 12\% of the video.
} \label{fig:task}
\vspace{-0.6cm}
\end{figure*}

Procedural videos illustrate how to perform a sequence of steps to complete diverse human tasks, such as cooking a meal or repairing a bike, and are driving a new frontier in AI coaching~\cite{huh2025vid2coach, grauman2024CVPRegoexo4d, panchal2024say} and robotics~\cite{bahlCVPR2023affordances, kareer2025ICRAegomimic, hoque2025egodex}. Given the routine and ubiquitous nature of procedural tasks, understanding procedural videos is essential for developing capable AI assistants, including those deployed on smart glasses that provide real-time contextual guidance 
across various domains such as sports, cooking and music~\cite{grauman2024CVPRegoexo4d, doughty2019pros, huang2024egoexolearn, wang2023ICCVholoassist, panchal2024say}, 
and dexterous robots designed to assist humans with similar tasks. Mistake detection---the task of identifying errors in the execution of a procedural activity---is particularly important, as it enables real-time support~\cite{panchal2024say}, facilitates skill assessment~\cite{parmar2019and, feng2025evostruggle, pan2025basket}, and helps identify areas for improvement during task execution~\cite{yi2025exact}. These capabilities make mistake detection a key component for personalizing learning and building more competent agents for procedural tasks in real-world environments.

Despite impressive progress, existing research on mistake detection largely operates in an offline, batch setting~\cite{ding2023every, Ghoddoosian2023CVPRata, Lee2024CVPRegoper, huang2025CVPRamnar, mazzamuto2025CVPRgazing}
that assumes access to the full video for identifying errors in the keysteps of a procedural activity. Consequently, such systems lack proactivity---that is, they cannot flag mistakes
before the agent has fully committed to or completed a step. This limitation reduces their applicability in real-time and interactive scenarios in which humans and robots operate, as well as in resource-efficient settings where avoiding wastage or damage to resources---such as ingredients during cooking or tools during a repair task---is critical. It also limits deployment in low-power setups~\cite{grauman2024CVPRegoexo4d, majumder2023chat2map, 9720526, 9933882}, 
\eg, on smart glasses, where the mistake detector must operate within an energy budget.

This motivates us to consider \emph{early} mistake detectors that operate on streaming procedural videos and identify errors in keysteps while observing only a minimal fraction of the video. %
Towards this goal, we introduce the task of \emph{early detection of mistakes in procedural videos}. In this task, 
the goal is to learn a policy 
that, given a streaming video of a keystep in a procedural activity---\eg, a chef \texttt{peeling a cucumber} for a salad, a mechanic \texttt{propping up a bike} to take off its wheel---decides \emph{how much} of the video to observe before \emph{exiting} (stopping inference).
The objective is to ensure that a mistake detector, conditioned on the recently observed frames, can accurately determine whether the keystep is a mistake or not while minimizing the fraction of the video that must be observed. See Fig.~\ref{fig:task}. 

We consider diverse categories of procedural mistakes in our task setting, including ordering errors (\eg adding oil to a saucepan before heating), technical mistakes 
(\eg over-tightening a bolt leading to threading damage),
and incorrect object usage (\eg using a spoon instead of a whisk to beat eggs).

To tackle this task, we propose MistExit, an early mistake detection framework comprising a mistake detector and a reinforcement learning (RL) policy. At each timestep of the streaming video, the detector processes the recently observed frames to produce an initial estimate of the keystep’s mistake label, along with predicted visual features for a sequence of future frames. Future anticipation enables the detector to better model the keystep’s correctness and provide reliable initial mistake estimates to the policy. The policy takes the detector’s outputs and the observed frames as inputs, aggregates them over time, and sequentially decides when to exit, while also producing the final prediction of the keystep’s correctness at the exit point. Whereas the detector outputs convey the detector’s uncertainty to the policy and indicate whether they are reliable enough to be incorporated into the policy’s model of the keystep’s correctness, the visual inputs---accumulated over time---allow the policy to verify the visual evidence against the detector’s predictions and adjust its exit strategy accordingly. 

We train our policy using a novel reward comprising three components: a dense term that promotes continuous improvement in detection quality and stabilizes training on longer and more complex videos; a sparse term that encourages accurate predictions at the exit point, capturing the essence of the task; and a video-aware time penalty that incentivizes early exits commensurate with the video’s length and complexity. 

We evaluate our approach on two real-world datasets of procedural videos: CaptainCook4D (CC4D)~\cite{peddi2024NeurIPScaptaincook4d} and Assembly101~\cite{sener2022CVPRassembly101}. Together, these datasets span diverse activity scenarios---cooking in CC4D vs. (dis)assembly in Assembly101---as well as different task environments, with an in-the-wild cooking setup in CC4D and a tabletop setting in Assembly101. Our model successfully learns to detect procedural mistakes accurately while exiting early, outperforming multiple strong baselines, including policies from early action recognition~\cite{wang2022CVPRadafocusV2, ghodrati2021CVPRframeexit, wu2019CVPRadaframe, fan2018IJCAIfastforward} and 
its variant
without future anticipation. 
\SM{We will release our code and data.}

%% file: sections/related_work.tex
\section{Related Work}\label{sec:related}

\paragraph{Mistake detection in \SM{procedural} videos.
}
Mistake detection in 
\SM{procedural} videos 
involves identifying deviations in the execution of keysteps within the demonstrated procedural activity~\cite{wang2023ICCVholoassist, sener2022CVPRassembly101, peddi2024NeurIPScaptaincook4d, grauman2024CVPRegoexo4d, Ghoddoosian2023CVPRata, flaborea2024CVPRprego, huang2025CVPRamnar, qian2022CVPRsvip}. Such deviations may include missing, adding or modifying certain steps~\cite{peddi2024NeurIPScaptaincook4d, Lee2024CVPRegoper, jang2019ICCVWepictent}, incorrectly performing some steps~\cite{wang2023ICCVholoassist, peddi2024NeurIPScaptaincook4d, haneji2024egooops, schoonbeek2024WACVindustreal, jang2019ICCVWepictent} or altering their order~\cite{sener2022CVPRassembly101, peddi2024NeurIPScaptaincook4d, jang2019ICCVWepictent}. Prior works identify mistakes of varying granularity--from sequence-level deviations~\cite{qian2022CVPRsvip, Ghoddoosian2023CVPRata} to errors in both fine-grained~\cite{wang2023ICCVholoassist, schoonbeek2024WACVindustreal, Lee2024CVPRegoper} and coarse-grained~\cite{sener2022CVPRassembly101, peddi2024NeurIPScaptaincook4d} actions, and tackle diverse activity scenarios--from indoor activities like cooking~\cite{peddi2024NeurIPScaptaincook4d, Lee2024CVPRegoper, grauman2024CVPRegoexo4d} and (dis-)assembly~\cite{sener2022CVPRassembly101, wang2023ICCVholoassist, qian2022CVPRsvip, grauman2024CVPRegoexo4d, schoonbeek2024WACVindustreal, Ghoddoosian2023CVPRata} to outdoor activities like pitching a tent~\cite{jang2019ICCVWepictent}. 

Most of these models operate in an \emph{offline} setting, assuming 
access to the entire video and identifying mistakes after observing it in full.  Such models determine step correctness by leveraging knowledge graphs constructed from video transcripts~\cite{ding2023every}, employing hand-crafted error functions that rely on action preconditions to decide step correctness~\cite{Ghoddoosian2023CVPRata}, comparing visual features of a step against pre-computed action-specific clusters~\cite{Lee2024CVPRegoper}, matching recognized step types against expected types extracted from a task graph~\cite{huang2025CVPRamnar}, or detecting inconsistencies between predicted and observed gaze trajectories of the actor~\cite{mazzamuto2025CVPRgazing}. 
In contrast, we address \emph{online} mistake detection—specifically, a challenging setting in which the goal is to determine whether a step is correct or erroneous as early as possible. 

\SM{Select prior methods tackle the challenging online (streaming) setting, but they detect only the first incorrect step using one-class classification to compare actions recognized by an action recognition model against expected actions derived from a large language model~\cite{flaborea2024CVPRprego, plini2024ti} or a procedural task graph~\cite{seminara2024differentiable}. However, these methods cannot detect mistakes beyond the first one because the action anticipation model becomes unreliable once the mistaken action is used as input for predicting subsequent actions.}
Unlike 
these models,
which can detect only the first mistake in a procedural sequence, we aim to identify all mistakes and do so early within each step. 
Finally, unlike all prior approaches, we equip our detector with the ability to anticipate future visual features, allowing it to flag mistakes even earlier and improve task performance, as we show in results.

\paragraph{Online video understanding.}
Prior works have studied various online video understanding tasks. Whereas online action recognition~\cite{bloom2014ICPRcluster, kviatkovsky2014CVIUcovariance, suarez2021AAAIonline} requires classifying the action in a video clip using a small fraction of frames, the goal in online action detection~\cite{de2016ECCVonline, eun2020CVPRlearning, gao2021CVPRwoad, wang2023ICCVmemory, wang2021ICCVoadtr, xu2019ICCVtemporal, zhao2022real} is to predict if a frame in a streaming video is from an action class or the background class. Other efforts tackle online action localization~\cite{soomro2016CVPRonline, kang2021ICCVcontext, kim2022ECCVslide} and estimate action start and end times, predict action segments in streaming videos~\cite{Shen2024CVPRprotas, ghoddoosian2022CVPRweakly} or estimate the continuous progress of an action in a streaming setting~\cite{Shen2024CVPRprotas, donahue2024CVPRself}. In contrast, we tackle a distinct task of detecting mistakes in 
\SM{procedural} videos in a streaming setting. 
Specifically, our focus is on both mistake detection accuracy and efficiency, and therefore, we evaluate an early detection setup where the goal is to recognize mistakes in a keystep as quickly as possible.

\paragraph{Early recognition in videos.}
Early recognition~\cite{ryoo2011ICCVearly, cao2013CVPRpartial, lan2014ECCVhiera, wang2020CVPRactive, chi2024infogcn++, wang2019CVPRprogressive} involves recognizing actions and events in videos by observing only a few frames at their onset. 
Earlier approaches
leverage probabilistic modeling~\cite{cao2013CVPRpartial, lan2014ECCVhiera, li2012ECCVaction, ryoo2011ICCVearly} by extracting hand-crafted features from partially observed videos. These methods employ techniques such as bag-of-words representations~\cite{ryoo2011ICCVearly}, sparse coding~\cite{kong2016TPAMImax, kong2014ECCVdisc}, or hierarchical representations~\cite{lan2014ECCVhiera}.
More recent methods use learned features~\cite{fernando2021CVPRanti, hu2019TPAMIsoft, wang2019CVPRprogress, zhaoICCV2019residual} produced through knowledge distillation between teacher models trained on full videos and student models trained on partial videos~\cite{cai2018AAAItransfer, fernando2021CVPRanti, hu2019TPAMIsoft, wang2019CVPRprogress}, generating full videos using partially observed ones~\cite{xu2019ACMMMcgan}, residual propagation~\cite{zhaoICCV2019residual}, inferring relations with graph neural nets~\cite{wu2021IJCVSpatialTemporalRR, wu2021AAAIAnticipatingFR, chi2024infogcn++}, learning both instance-specific features and generic features shared across the dataset~\cite{foo2022ECCVera} or multi-scale representation of partial videos~\cite{stergiou2023CVPRwisdom}. In contrast, we tackle the novel task of early recognition of mistakes 
in \SM{procedural} videos.
Moreover, unlike the above methods that rely on short, fixed-length video segments, our goal is to learn a model that \emph{adaptively} determines an early exit point aligned with the video’s length and complexity.

More closely related to our work are methods that learn early exit policies for adaptively stopping inference in a streaming video when the recognition backbone makes a correct prediction, thereby aiming to achieve high accuracy while saving on inference cost~\cite{wang2022CVPRadafocusV2, wang2022ECCVadafocusv3, wang2024TPAMIuni, ghodrati2021CVPRframeexit, wu2019CVPRadaframe, fan2018IJCAIfastforward}. While some methods~\cite{wang2022CVPRadafocusV2, wang2022ECCVadafocusv3} use simple heuristics like making the exit decision by comparing the prediction confidence~\cite{wang2022CVPRadafocusV2} or prediction entropy~\cite{wang2022ECCVadafocusv3, wang2024TPAMIuni} against a pre-defined threshold,
FrameExit~\cite{ghodrati2021CVPRframeexit}
trains an exit policy in a supervised manner using exit pseudo-labels generated by comparing the recognition loss against video-progress-dependent thresholds during training. Some approaches~\cite{wu2019CVPRadaframe, fan2018IJCAIfastforward} also use reinforcement learning (RL) to train exit policies with both dense rewards~\cite{wu2019CVPRadaframe} that encourage an exit when there is no further potential for improvement in the prediction confidence for the target class, and sparse rewards~\cite{fan2018IJCAIfastforward} that not only incentivize maximizing the target prediction confidence at the point of exit but also punish the model for consuming too many frames.  

Different from all early exit methods, we focus on achieving early exit for mistake detection 
in \SM{procedural} videos 
and we do so by learning an RL-based exit policy. 
\SM{Our policy design enables the model to continuously refine its estimate of a keystep’s correctness by temporally aggregating visual information together with initial mistake estimates from the detector. Combined with an RL reward that promotes improving detection quality over time while encouraging early and accurate exits, this design leads to more stable training on longer and more complex episodes and substantially outperforms existing RL-based alternatives.
}

\vspace{-0.3cm}

%% file: sections/task.tex
\section{Early mistake detection task}\label{sec:task}
We introduce a new task: early detection of mistakes in 
\SM{procedural} videos.
In this task, the goal is to train a policy that operates on a streaming video of a keystep in 
\SM{a procedural} activity
(\eg, a chef \texttt{whisking eggs} when preparing an omelette, a mechanic \texttt{jacking up a car} when fixing a flat tire) and intelligently decides \emph{how much} of the video to observe before exiting \SM{(stopping inference)}, such that the observed length is \emph{very low} compared to the full length of the video, while also ensuring a mistake detection model 
\emph{accurately} detects any mistakes in the video.
\SM{Using visual inputs, potentially alongside the mistake detector’s predictions, the policy must 
infer whether
 sufficient evidence has been gathered to determine the step’s correctness and accordingly decide whether it is 
 a good time to exit.
 }

\vspace{-0.2cm}

\paragraph{Task definition.} 
Formally, we consider 
\SM{a procedural} video
$V$ in a streaming setting. $V$ consists of a sequence of $N$ video clips\footnote{\SM{We assume pre-segmented keysteps, as modern action detection models~\cite{wang2021ICCVoadtr, an2023ICCVmini, zhao2022real} can robustly estimate keystep start and end times}.}, such that $V$ = $[V_1, \ldots, V_N]$. Each clip $V_i$ corresponds to a keystep in 
the \SM{procedural} activity
shown in $V$ and has length $T_i$. 
Clip $V_i$ streams RGB frames 
at the rate of $f$ frames per second, such that $I^{\hat{t}}_i$ denotes the image frame at time $t$ and $\hat{t} = \lfloor t * f \rfloor$ provides the frame index for time $t$. Here, $0 \leq t \leq T_i$ and $\lfloor y\rfloor$ indicates rounding down $y$.

In this task, we aim to train a model 
comprising a mistake detector $\mc{D}$ and an exit policy $\pi$. Given a streaming clip $V_i$, the policy $\pi$ must take an action $a^t_i \in \mathcal{A}$ at each time $t$, 
where \SM{$\mathcal{A} = \{Continue, Exit\text{-}mistake, Exit\text{-}correct\}$} denotes the action space. 
$V_i$ keeps producing image frames until the policy decides to $Exit$\SM{---where $Exit\text{-}mistake$ and $Exit\text{-}correct$ correspond to stopping inference with the model’s final prediction for the clip being \emph{mistake} or \emph{correct}, respectively---}or until $V_i$ reaches its end. 

Let $E_i$ denote this end point---natural or policy-induced---of the image stream. At any time $t$ before the end point $E_i$, the mistake detector $\mc{D}$ takes the sequence of $K$ most recent frames,
$S^t_i$, such that $S^t_i = [I^{\hat{t}-K+1}_i, \ldots, I^{\hat{t}}_i]$, and produces an estimate $\tilde{M}^t_i$ of the mistake label $M_i$---0 for a mistake and 1 for correct---for $V_i$. $M_i$ indicates if the keystep corresponding to $V_i$ is being done correctly or not, 
\SM{and provides the policy with an initial estimate of the step's correctness}. 
\SM{Given the visual frames and the mistake detector's predictions,} the policy 
\SM{must} intelligently decide 
when to exit, such that the resulting end point $E_i$ occurs very early vis-a-vis the full clip length, \ie $E_i \ll T_i$, 
\SM{while its final prediction, as extracted from its action at the end point, matches the mistake label. That is, $\mathbbm{\mathcal{E}}(a^E_i) = M_i$, where $\mathcal{E}$ extracts the clip’s predicted correctness type from an $Exit$ action.}

Succeeding at this task requires a strong synergy between the exit policy and the mistake detector. The policy must reason from visual inputs\SM{---potentially alongside the detector’s predictions---about how far}
the keystep has progressed and whether sufficient evidence has been gathered for reliable mistake detection. 
In turn, the mistake detector must produce accurate estimates early in the streaming clip \SM{so that the policy can make prompt exit decisions}.
\SM{We evaluate the resulting performance–efficiency trade-off by measuring both the detection accuracy at the exit point and the fraction of the clip consumed (cf.~Sec.~\ref{sec:exp_setup}).}

\vspace{-0.3cm}

%% file: sections/approach.tex
\section{Approach}\label{sec:approach}
\begin{figure*}[!t] 
    \centering
    \includegraphics[width=0.9\linewidth]{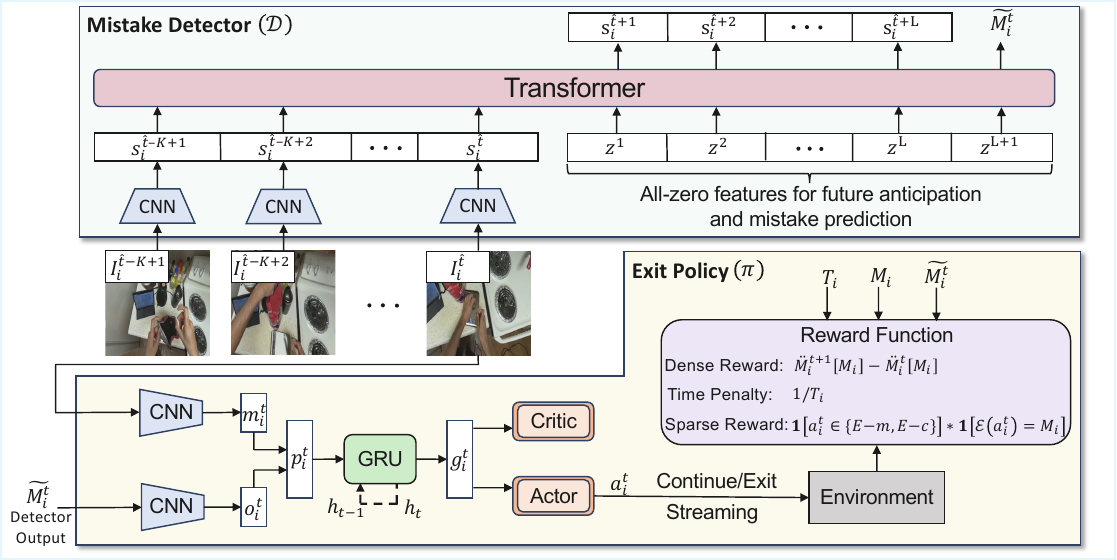} 
    \vspace*{-0.1in}
    \caption{Our MistExit model for early mistake detection has two components: 1) a mistake detector $\mc{D}$ (top), and 2) an exit policy $\pi$ (bottom).  At each timestep in a streaming keystep clip, $\mc{D}$ processes recent frames to predict the keystep’s mistake label and anticipate future features, improving the mistake detection quality. The policy $\pi$ takes the detector’s estimate and the latest frame, aggregates them over time, and decides when to exit. We train $\pi$ with a novel reward that encourages improving detection quality over time while promoting early and accurate exits.
    } \label{fig:model}
    \vspace{-0.5cm}
\end{figure*}

We pose our early mistake detection task as a reinforcement learning (RL) problem, and 
propose our \textbf{MistExit} model to solve the task. Our model has two key components: \textbf{1)} a mistake detector $\mc{D}$, and \textbf{2)} an exit policy $\pi$. 
Given a streaming clip corresponding to a keystep in an instructional video, the mistake detector determines whether the step is correct or erroneous by \SM{not only} leveraging recently observed frames \SM{to detect mistakes in the parts of the clip observed so far} 
\SM{but also} anticipating future visual features.
\SM{Anticipation lets the model infer if the step can lead to a mistake in the \emph{future} (\eg, vigorously beating eggs in a small bowl may indicate a future spillover, taking off a tire before jacking up the car may damage the wheel hub), so that it can better facilitate early exits.}

Conditioned on the detector’s latest prediction and the current frame 
\SM{from the stream---aggregated over time---the policy}
sequentially decides whether to continue observing or to exit.
\SM{On the one hand,} the detector’s confidence scores provide an explicit signal of \SM{its current prediction} 
uncertainty \SM{to the policy}, guiding \SM{it}
to either gather additional evidence or rely on accumulated predictions \SM{from the past} 
for immediate termination. \SM{On the other hand,} 
the visual frames \SM{observed so far}, together with the detector's \SM{initial estimates},  
\SM{enable} the policy
\SM{to cross-check the visual evidence with the detector outputs and assess whether the detector’s estimates are reliable enough to be used 
in the decision making, especially when the estimtes are high-confidence.}
See Fig.~\ref{fig:model}. Next, we describe these two components in detail. 
\vspace{-0.3cm}

\subsection{Mistake detector $\mc{D}$}\label{sec:detector}
\SM{Our mistake detector $\mc{D}$ comprises a transformer encoder~\cite{vaswani2017NIPSattn, flaborea2024CVPRprego} that takes as input a sequence of the most recently observed frames and predicts both an estimate of the keystep's binary mistake label 
and the visual features for a sequence of future frames.}
\SM{By predicting the features for future frames, $\mc{D}$ learns to anticipate the future and consequently can better model if a keystep is likely to end up being a mistake or not. This helps the policy to make even earlier exits while also ensuring improved mistake detection accuracy, as we show in results.}

Specifically, we first use 
a CNN encoder~\cite{carreira2017CVPRi3d, lin2019ICCVtsm}
to project the frame sequence $S^t_i$ (cf.~Sec.~\ref{sec:task}) into a visual feature sequence $s^t_i$, such that $s^t_i = [s^{\hat{t} - K + 1}_i, \ldots, s^{\hat{t}}_i]$ (cf.~Sec.~\ref{sec:task}).
Next, we produce a sequence of $L + 1$ all-zero features, $z$, such that 
first $L$ entries correspond to the features for the
future frames that $\mc{D}$ anticipates,
and the last feature is the feature whose output will be used for estimating the mistake label. We then concatenate $s^t_i$ and $z$ into a single feature sequence $d^t_i = [s^t_i, z]$. 
We further add to each feature entry in $d^t_i$ the appropriate sinusoidal positional embeddings~\cite{vaswani2017NIPSattn} and a learnable modality embedding~\cite{majumder2022NEURIPSfew, majumder2025ICCVswitch} to distinguish among three different feature types: %
one for all features in $s^t_i$, one for the first $L$ features in $z$, corresponding to the anticipated features, and one for the last feature in $z$, which corresponds to the feature that is mapped to the mistake logits. 

Next, we aggregate $d^t_i$ using transformer encoder layers~\cite{vaswani2017NIPSattn, flaborea2024CVPRprego, zhang2022CVPRaction, Lee2024CVPRegoper} and store the last $L+1$ output features in a sequence $\bar{d}^t_i$. Finally, we transform the first $L$ entries and the last entry in $\bar{d}^t_i$ using separate MLPs to obtain the estimates $\tilde{s}^t_i$ for the anticipated features, $\bar{s}^t_i = [s^{\hat{t} + 1}_i, \ldots, s^{\hat{t} + L}_i]$, and the logits $\tilde{M}^t_i$ for \SM{binary} mistake classification.
\vspace{-0.3cm}

\subsection{Exit policy $\pi$}\label{sec:policy}
Our second model component is our exit policy $\pi$ that takes the latest visual frames from the streaming clip and the predictions from the mistake detector\SM{, aggregates these inputs over time,} and actively %
decides when to stop observing further frames. Specifically, for a clip $V_i$, it uses the visual cues and detection scores to predict a sequence of actions $a^t_i$ that leads to an accurate prediction of the keystep's correctness at the exit point $E_i$, while keeping the value of $E_i$ as low as possible (cf.~Sec.~\ref{sec:task}). The policy comprises two main modules: 1) an input encoder, and 2) a policy network.

\vspace{-0.2cm}

\paragraph{Inputs and encoding.}
At every time $t$ in a streaming video clip $V_i$, the exit policy receives a visual observation  $O^t_i$, along with the latest predictions $\tilde{M}^t_i$ from the mistake detector. $O^t_i$ comprises the latest frame $I^{\hat{t}}_i$ from the stream (cf.~Sec.~\ref{sec:task}). 

The visual input $O^t_i$\SM{, coupled with cues from the temporally aggregated past inputs $O^{<t}_i$,}
enables the model to infer the current stage of the activity—early or late—and to determine whether the frames observed thus far provide sufficient information to assess the correctness of the keystep. 
The mistake detection scores $\tilde{M}^t_i$ directly inform the model of the detector's prediction and 
its confidence
at the present time. 
\SM{The policy accumulates these estimates along with visual evidence over time and determines whether sufficient information has been gathered to decide the step’s correctness, adjusting its exit point accordingly.}

To encode the visual observation $O^t_i$, we use a CNN~\cite{carreira2017CVPRi3d, lin2019ICCVtsm}
encoder and produce 
a 1D embedding $o^t_i$. For the \SM{binary} mistake detection scores $\tilde{M}^t_i$, we first normalize them \SM{and obtain detection confidences} by passing through a softmax layer 
and then encode 
\SM{the confidences $\ddot{M}^t_i$} using
an MLP encoder 
to produce another 1D embedding $m^t_i$. Next, we concatenate $o^t_i$ and $m^t_i$ along the channel dimension to produce our policy embedding $p^t_i$, such that $p^t_i = [o^t_i, m^t_i]$.

\paragraph{Policy network.}
The policy network begins with a gated recurrent unit (GRU)~\cite{cho2014EMNLPgru, chung2014NIPSWgru} that uses the policy embedding $p^t_i$ and the policy's aggregated history of states $h^{t-1}_i$ to produce an updated history $h^t_i$ and a representation of the current state, $g^t_i$. Next, an actor-critic module takes $g^t_i$ and $h^{t-1}_i$ as inputs, and generates the policy distribution $\pi_{\theta}(a^t_i | g^t_i, h^{t-1}_i)$ and the value of the state, $\mathcal{V}_{\theta}(g^t_i, h^{t-1}_i)$, where $\theta$ denotes the policy parameters. 
Finally, the policy samples an action $a^t_i$ from the policy distribution $\pi_{\theta}$, thereby deciding to $Continue$ (cf.~Sec.~\ref{sec:task}) consuming frames or stop further inference and $Exit$ the stream. 
\SM{If the policy decides to exit, the model’s final prediction is determined by the chosen exit action--correct for $Exit\text{-}correct$ and mistake for $Exit\text{-}mistake$.}
\SM{Importantly, the two $Exit$ actions enables the policy to temporally aggregate the detector’s initial predictions, which might be noisy due to the fixed-length observation window,
and generate a more reliable and accurate final prediction at the exit point, guided by our reward function (Sec.~\ref{sec:train}).}

\vspace{-0.3cm}

\subsection{Model training}\label{sec:train}
\paragraph{Mistake detector training.}
We set the training loss $\mc{L}^{\mc{D}}$ for our mistake detector $\mc{D}$ to a weighted sum of the mistake detection loss $\mc{L}^M$ and a 
\SM{future anticipation loss $\mc{L}^F$}, such that
\begin{align}
        \mc{L}^{\mc{D}} = w_1 * \mc{L}^M + w_2 * \mc{L}^F.
\end{align}

Here, $\mc{L}^M$ is the cross-entropy loss ($CE$) between the mistake label $M_i$ (cf.~Sec.~\ref{sec:detector}) and the detector's estimate of the same, $\tilde{M}^t_i$ (cf.~Sec.~\ref{sec:policy}), such that
\begin{align}
    \mc{L}^M = CE(M_i, \tilde{M}^t_i),
\end{align}
\SM{Our future anticipation loss $\mc{L}^F$} is the average L1 loss 
between the ground-truth future feature sequence $\bar{s}^t_i$ and its estimate $\tilde{s}^t_i$, such that 
\begin{align}
\mc{L}^F = \frac{1}{L} * (||s^{\hat{t} + 1}_i - \tilde{s}^{\hat{t} + 1}_i||_1 + \ldots + ||s^{\hat{t} + L}_i - \tilde{s}^{\hat{t} + L}_i||_1),
\end{align}
and $w_1$ and $w_2$ are the weights on the two losses, respectively. We set $w_1$ and $w_2$ on the basis of a disjoint validation split. 

\paragraph{Policy training.}
We propose a novel reward function to train our exit policy $\pi$:
\begin{align}
\vspace{-0.3cm}
    r^t_i = 
          \begin{cases}
            v_1 * \big(\underbrace{\ddot{M}^{t+1}_i[M_i] - \ddot{M}^t_i[M_i]}_{\scriptscriptstyle
 \text{Dense reward}}\big) - v_3 * \big(\underbrace{1 / T_i}_{\scriptscriptstyle
 \text{Penalty}}\big) &\quad t \in [0, E_i - 1]\\
            v_2 * \underbrace{\mathbbm{1}[a^t_i \in \{E\text{-}m, E\text{-}c\}] * \mathbbm{1}[\mathcal{E}(a^t_i) = M_i]}_{\scriptscriptstyle \text{Sparse reward}}  &\quad t = E_i.\\ 
        \end{cases}
\vspace{-0.3cm}
\end{align}
Here, $\ddot{M}^t_i$ denotes the detector’s predicted confidence scores obtained by applying a softmax normalization to $\tilde{M}^t_i$ (cf.~Sec.~\ref{sec:detector}). 
$M_i$ represents the ground-truth mistake label for the keystep (cf.~Sec.~\ref{sec:task}). 
We use $E\text{-}m$ and $E\text{-}c$ to denote the $Exit\text{-}mistake$ and $Exit\text{-}correct$ actions, respectively. 
Finally, $\mathcal{E}$ extracts the clip’s predicted correctness type from the policy’s action at the exit time. 
\SM{In case the policy doesn't exit, we use the detector's final prediction as the mistake estimate.}

Our reward has three terms: 1) a dense reward, 2) a sparse reward, and 3) a penalty. We set the dense reward to the improvement in the 
prediction confidence $\ddot{M}_i[M_i]$ for the mistake label 
from time $t$ to $t+1$, thereby incentivizing the policy to let the detector continuously improve its estimate of the step's correctness over time.
\SM{Moreover, our dense reward better handles long-horizon episodes arising from long and complex keysteps, improving RL training convergence.}

We additionally provide the policy with a sparse reward when it chooses to exit and its predicted clip type—derived from its action—matches the ground-truth label. This encourages the policy to refine potentially unreliable intermediate estimates from the mistake detector and produce a high-quality prediction at the exit point.
Our last term is a time penalty that discourages late exits by the policy. Importantly, instead of using a fixed penalty for all clips, we set it to the inverse of the clip length $T_i$ 
(cf.~Sec.~\ref{sec:task}). This allows the detector to consume more frames for determining the step's correctness if the clip is longer and hence, possibly more challenging. Finally, we weight the sparse reward, the dense reward and the time penalty with $v_1$, $v_2$ and $v_3$, respectively, where the weights are chosen using a held-out validation set.

\paragraph{Training curriculum.}
We pre-train our mistake detector $\mc{D}$ without an exit policy by first randomly selecting a clip $V_i$ and then randomly sampling a time $t$ in the clip. 
Next, we train our policy $\pi$ while freezing the pre-trained detector. 
\SM{Since our reward 
depends on
the mistake detector outputs, freezing the detector during policy training ensures 
stationary rewards
and improves convergence.}

\vspace{-0.2cm}

%% file: sections/experiments.tex
\section{Experiments}
\label{sec:experiments}
Here, we give an overview of setup details and then provide results. 

\vspace{-0.2cm}

\subsection{Experimental setup}\label{sec:exp_setup}
\paragraph{Datasets.}
We evaluate our model on two instructional video datasets: CaptainCook4D (CC4D)~\cite{peddi2024NeurIPScaptaincook4d} and Assembly101~\cite{sener2022CVPRassembly101}. Whereas CC4D comprises cooking videos recorded in real-world kitchens, Assembly101 contains videos of assembling and disassembling toy vehicles. The two datasets lets us evaluate diverse activity scenarios. In both datasets, each video consists of a sequence of keysteps, where each keystep may be performed either correctly or incorrectly. While both datasets provide two high-level mistake labels—\emph{correct} and \emph{mistake}, Assembly101 includes an additional category in its mistake taxonomy, \emph{correction}, which we treat as a mistake in our experiments. The video segments corresponding to individual keysteps serve as the streaming clips in our task (cf.~Sec.~\ref{sec:task}). We construct train/val/test splits containing 2624/200/1037 
clips for CC4D and 2726/200/343 clips for Assembly101, respectively, ensuring that clips across different splits are drawn from disjoint videos.
 This amounts to a total of 58.3 and 14.5 hours of video for CC4D and Assembly101, respectively. We set the frame rate to 2 fps for all videos.

\paragraph{Implementation.}
We modify the action recognition module of the state-of-the-art PREGO~\cite{flaborea2024CVPRprego} model to enable future anticipation and use it as our mistake detector. Specifically, given a streaming clip and any point in time in the clip, our mistake detector $D$ takes as inputs the visual features for the $K=5$ most recent frames and outputs an estimate of the features for the next $L=20$ frames, in addition to predicting if the clip is a mistake or not. We train the detector until convergence by setting the weights in its training loss (cf.~Sec.~\ref{sec:train}) to $w_1=1$ and $w_2=10^{-1}$, and using the AdamW~\cite{loshchilov2017ArXiVadamw} optimizer with a batch size of 128, an initial learning rate of $10^{-6}$, and a  weight decay of $5 \times 10^{-2}$. 
We use PPO~\cite{schulman2017ArXiVppo} to train our exit policy $\pi$ for a total of 42 million policy steps with 4 PPO updates after every 40 steps. To this end, we set the weights of our reward components (cf.~Sec.~\ref{sec:train}) to $v_1=10^{-1}$ and $v_2 = v_3 = 1$ and use the Adam~\cite{kingma2014ArXiVadam} optimizer with a batch size of 8 and an initial learning rate of $10^{-4}$. See Supp. (Sec.~\ref{sec:supp_implementation}) for more details.

\paragraph{Baselines.}
We compare against the following baselines and SOTA methods:
\begin{itemize}[leftmargin=*,topsep=0pt,partopsep=0pt,itemsep=0pt,parsep=0pt]
    \item \textbf{Random:} a policy that randomly chooses 
    \SM{an action from action space $\mc{A}$}
    \item 
\textbf{AdaFocusV2~\cite{wang2022CVPRadafocusV2}:} 
    a policy that exits when the confidence of the mistake detector’s predicted class exceeds a threshold of 0.75. \SM{We also evaluate an enhanced version of this model, AdaFocusV2++, where the exit decision uses the mean confidence over the past $P$ predictions, with $P=5$ for CC4D and $P=3$ for Assembly101.}
    \item \textbf{AdaFocusV3~\cite{wang2022ECCVadafocusv3}:} 
    policy that exits when the entropy of the mistake detector’s predictions is below 0.1 for CC4D and 0.5 for Assembly101. \SM{Similar to AdaFocusV2++, we also evaluate AdaFocusV3++, which computes the mean entropy over the past $P$ predictions, with $P=3$ for CC4D and $P=5$ for Assembly101.}
    \item \textbf{FrameExit~\cite{ghodrati2021CVPRframeexit}:} a \SM{learned} baseline that produces exit pseudo-labels by checking the mistake detector's prediction loss against a time-dependent threshold---lower for earlier frames and higher for later ones---and trains a policy via supervised learning using these pseudo-labels.
    \item \textbf{FastForward~\cite{fan2018IJCAIfastforward}:} \SM{another learned} baseline that trains an RL policy using a reward that encourages the detector to continuously improve its predicted confidence of the target class over time while simultaneously discouraging late exits through a fixed penalty of $10^{-2}$ per frame consumed. 
    \item \textbf{AdaFrame~\cite{wu2019CVPRadaframe}:} an RL baseline that trains an actor-critic network with a dense reward defined as the improvement, if any, in the detector’s predicted confidence for the target class relative to its previous best confidence, and stops inference when the value estimate produced by the critic falls below the previous maximum predicted value by at least 0.7 twice in an episode.
\end{itemize}

For fair comparison, all 
models use \SM{the same mistake detection backbone~\cite{flaborea2024CVPRprego} and are trained and validated using our train and val splits}. 
\SM{Importantly, all three learned models were originally designed for skip-forward/backward video recognition with \emph{offline} video access; we adapt them to our streaming setup by equipping them with our action space.}
We select policy-specific hyperparameters—such as the threshold for the 
AdaFocus family 
and the time penalty for FastForward—based on validation performance.

\paragraph{Evaluation metrics.}
\SM{Following existing early recognition works~\cite{ryoo2011ICCVearly, cao2013CVPRpartial, lan2014ECCVhiera}}, we assess our model along two dimensions: \textbf{1)} mistake detection accuracy, and \textbf{2)} earliness of exit. We capture this accuracy vs. efficiency trade-off through scatter plots (shown below), where stronger models achieve higher accuracy while exiting earlier. To gauge accuracy, we use the \textbf{average precision} (AP) metric, which measures the area under the precision vs. recall curve. To measure model efficiency, we compute the mean 
\SM{\textbf{observation ratio} (OR)~\cite{ryoo2011ICCVearly, cao2013CVPRpartial, lan2014ECCVhiera}}
across all test clips, where each exit time is normalized by the corresponding clip length and then averaged over all clips.

\subsection{Early mistake detection results}\label{sec:results}

\begin{figure}[!th]
    \centering
    \begin{subfigure}[b]{0.48\linewidth}
    \centering
    \includegraphics[width=\linewidth]{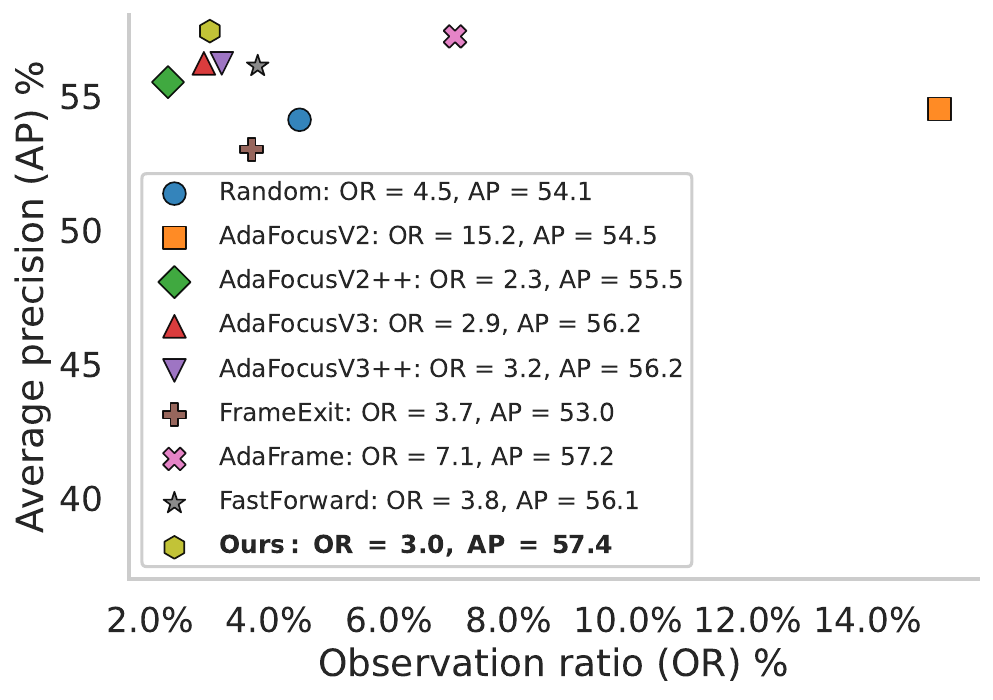}
    \caption{ CaptainCook4D~\cite{peddi2024NeurIPScaptaincook4d}
    }
    \label{fig:cc4d_main}
    \end{subfigure}\hfill
    \begin{subfigure}[b]{0.48\linewidth}
    \centering
    \includegraphics[width=\linewidth]{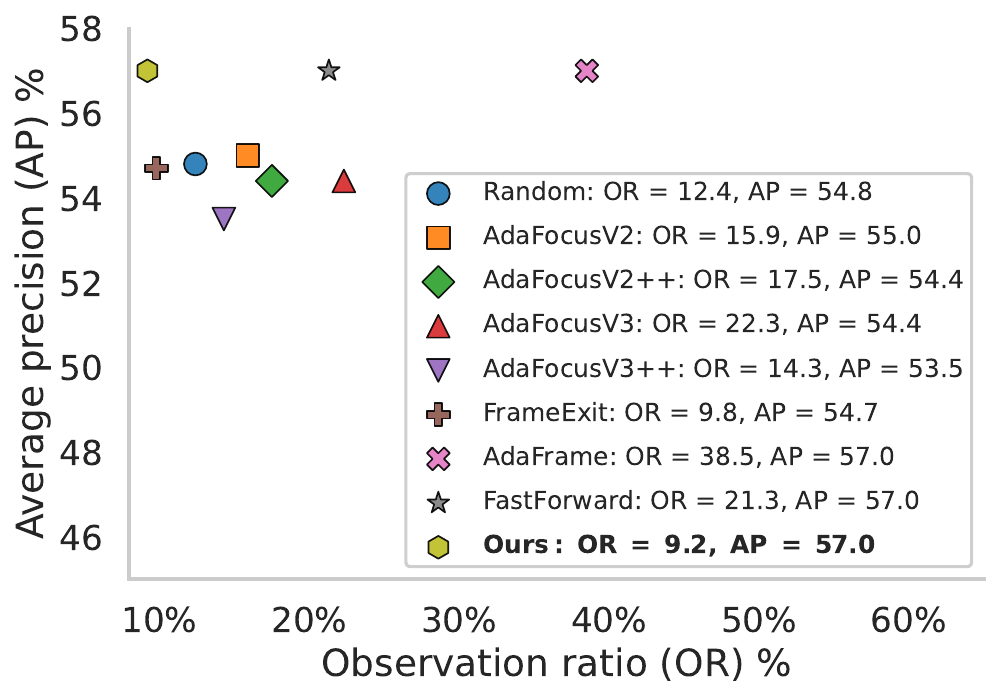}
    \caption{Assembly101~\cite{sener2022CVPRassembly101}
    }
    \label{fig:assembly_main}
    \end{subfigure}
\caption{
Early mistake detection results. 
Higher AP, lower OR is better. 
}
\vspace{-0.5cm}
    \label{fig:main_results}
\end{figure}

Fig.~\ref{fig:main_results} shows the early detection quality of all models on both CaptainCook4D (CC4D)~\cite{peddi2024NeurIPScaptaincook4d} (Fig.~\ref{fig:cc4d_main}) and Assembly101~\cite{sener2022CVPRassembly101} (Fig.~\ref{fig:assembly_main}). Employing a naive exit heuristic, such as Random, is insufficient for achieving high-quality and efficient mistake detection, highlighting the challenging nature of our early mistake detection task. In comparison, the AdaFocus~\cite{wang2022CVPRadafocusV2, wang2022ECCVadafocusv3} policy family, which leverages prediction uncertainty to guide exit decisions, improves detection accuracy while enabling earlier exits, particularly on CC4D. This suggests that the mistake detector’s predictions provide useful cues for determining the appropriate exit point. The FrameExit~\cite{ghodrati2021CVPRframeexit} further improves performance, especially on Assembly101, demonstrating the usefulness of learning an exit policy. Both RL baselines, AdaFrame~\cite{wu2019CVPRadaframe} and FastForward~\cite{fan2018IJCAIfastforward}, significantly improve detection accuracy compared to FrameExit, indicating that policies trained with RL, rather than supervised learning, can better adapt their exit decisions to the nature and complexity of the video. 
 
Our model outperforms all baselines on CC4D while 
also improving efficiency over most. 
On Assembly101, our model ranks again among the top performers while achieving significantly better efficiency. Notably, our gains over the other RL-based counterparts highlight the benefits of our superior reward formulation and the augmented input space, which includes the mistake detector’s prediction scores alongside the RGB frames. Moreover, comparing our performance with the FrameExit method illustrates that our well-designed RL policy not only substantially boosts the mistake detection accuracy but also enables earlier exits---identifying early segments of a streaming clip that are indicative of the keystep's correctness. Overall, our approach consistently delivers a stronger accuracy-efficiency trade-off than existing methods.

\subsection{Model analysis}\label{sec:model_analysis}
\begin{figure}[!th]
    \centering
    \begin{subfigure}[b]{0.48\linewidth}
    \centering
    \includegraphics[width=\linewidth]{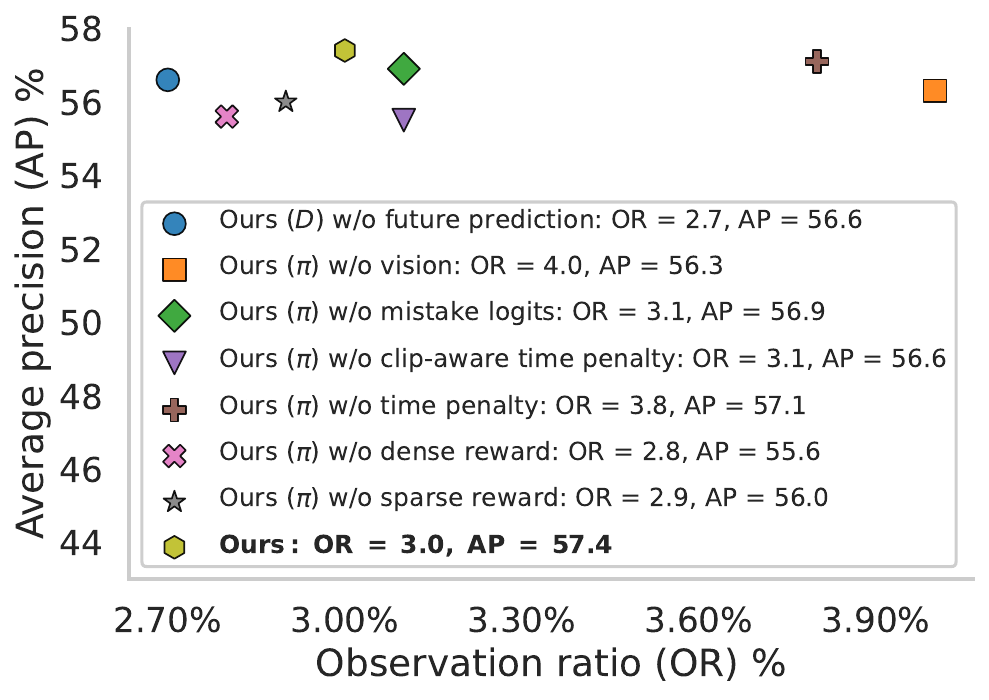}
    \caption{Model ablations}
    \label{fig:cc4d_abs}
    \end{subfigure}\hfill
    \begin{subfigure}[b]{0.48\linewidth}
    \centering
    \includegraphics[width=\linewidth]{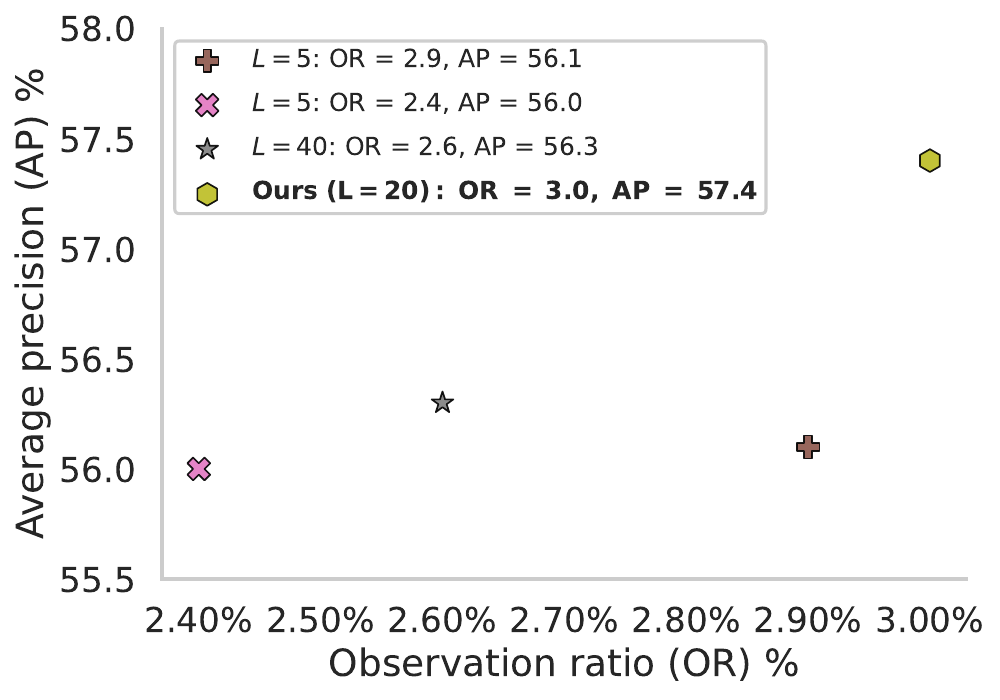}
    \caption{Performance vs. detector anticipation length}
    \label{fig:analysis_anticipationLength}
    \end{subfigure}
\caption{
\textbf{Left:} Ablation on larger-scale CaptainCook4D (CC4D)~\cite{peddi2024NeurIPScaptaincook4d}. 
\textbf{Right:} Early mistake detection results on larger-scale CC4D for different lengths ($L$) of the anticipated feature sequence in our mistake detector (Sec.~\ref{sec:detector}).
Higher AP, lower OR is better for both plots. 
}\label{fig:ablations_n_anticipatedFeatureLengthImpact}
\vspace{-0.5cm}
\end{figure}

\paragraph{Ablations.}
In Fig.~\ref{fig:cc4d_abs}, we show the results from ablating our model components. Disabling future anticipation in the mistake detector leads to a substantial drop in detection accuracy.
This indicates that future anticipation (one of our contributions) allows the model to delay 
exit slightly to locate informative clip segments that yield more reliable mistake predictions. 

Removing visual inputs from the policy drastically increases the observation ratio, highlighting the importance of visual information for determining when sufficient evidence has been gathered. In particular, visual cues allow the policy to reason about the stage of the activity and whether it is too early to exit, while also cross-checking the visual evidence with the detector’s predictions to assess whether the detector’s estimates are reliable enough to trigger an immediate exit. Removing access to the mistake detector’s logits also degrades performance, indicating that our model effectively leverages these initial estimates to build an implicit representation of the step’s correctness and adjust its exit point accordingly. Using a 

fixed time penalty instead of one conditioned on the clip length also reduces performance, underscoring its role in helping the policy determine an appropriate exit point based on the length and complexity of the clip. Removing the time penalty also degrades performance, particularly policy efficiency, highlighting its role in encouraging early exits. Furthermore, because our task involves long horizons arising from variable clip lengths, the dense reward is critical for stable RL training, without which, the detection accuracy drops significantly. Removing the dense reward also affects detection accuracy, as it provides an important learning signal that captures the core objective of our task—maximizing detection accuracy at the point of exit.

\paragraph{Analysis of length of anticipated frame sequence $L$.}

In Sec.~\ref{sec:experiments} in main, 
we evaluated our method with the length of the anticipated future feature sequence in the mistake detector 
(Sec.~\ref{sec:detector}) set to $L=20$ (Sec.~\ref{sec:exp_setup}). 
Here, we analyze the impact of $L$ on our model performance. Fig.~\ref{fig:analysis_anticipationLength} presents the results of this analysis. We observe that setting $L=20$ significantly improves mistake detection performance compared to variants with other values of $L$, while only marginally increasing the amount of video consumed on average. This suggests that with $L=20$, our model is better able to anticipate how the step will evolve and leverage this information to improve mistake detection accuracy, even though doing so may require slightly more visual evidence.

\paragraph{Qualitative examples.}

\begin{figure*}[t!] 
    \centering
    \includegraphics[width=1.0\linewidth]{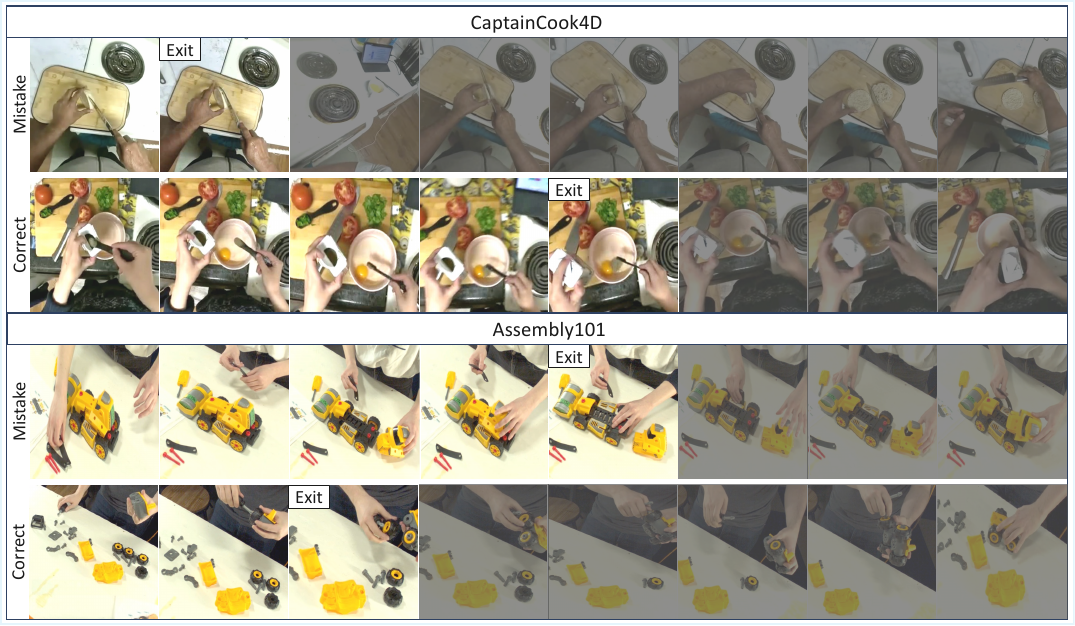} 
    \caption{
    Our model's successful predictions on CaptainCook4D~\cite{peddi2024NeurIPScaptaincook4d} (top 2 rows) and Assembly101~\cite{sener2022CVPRassembly101} (bottom 2 rows). Our model correctly detects mistakes by identifying cues such as incorrect technique---for example, the knife positioned to produce an abnormally thin slice in row 1---and signs of struggle by the actor, such as repeatedly moving the cabin back and forth in row 3. It can also correctly predict that a step will end in a successful execution by leveraging cues that indicate the correctness of the remaining portion of the step---for instance, a closed pepper container in row 2 may suggest that the actor will not add extra pepper to the eggs, while a correctly installed wheel in row 4 may indicate that the remaining wheels will also be installed correctly.
    } \label{fig:qual}
    \vspace{-0.5cm}
\end{figure*} 

In Fig.~\ref{fig:qual}, we show some success cases from CaptainCook4D~\cite{peddi2024NeurIPScaptaincook4d} (top 2 rows) and Assembly101~\cite{sener2022CVPRassembly101} (bottom 2 rows). Notably, our model can detect incorrect techniques and flag mistakes early by anticipating that the step will ultimately result in an error. \Eg, in row 1, the actor begins by holding the knife too close to the base of the bun, resulting in an abnormally thin cut, and our model successfully picks up on this cue. Furthermore, our model can detect signs of struggle and repeated actions by the actor, which often indicate that the step will be labeled as a mistake. \Eg, in row 3, the actor unsuccessfully attempts to attach the cabin to the chassis and moves it back and forth multiple times. Our model likely identifies this repeated motion and infers that the actor will fail to successfully complete the step.

\begin{figure}[!t] 
    \centering
    \includegraphics[width=1.0\linewidth]{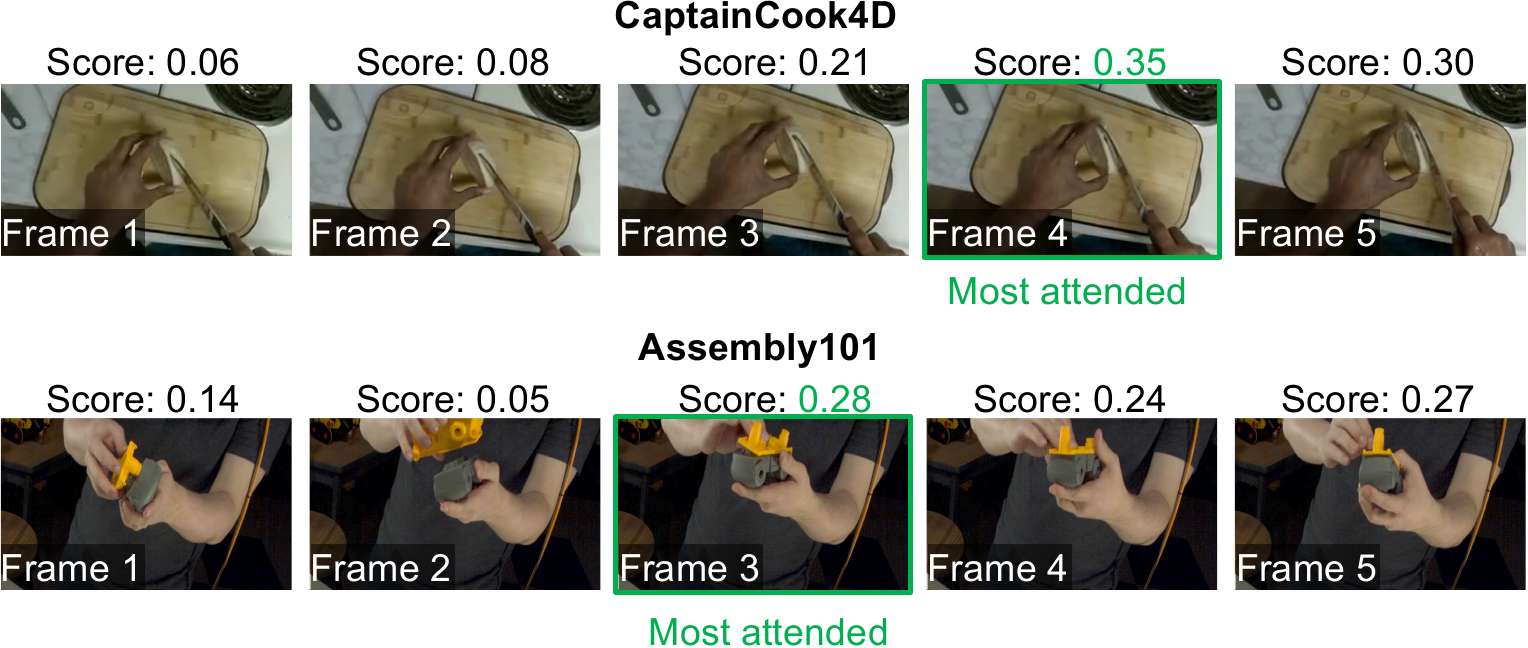} 
    \caption{Qualitative examples illustrating our model’s frame-level attention scores at the exit point for CaptainCook4D~\cite{peddi2024NeurIPScaptaincook4d} (top) and Assembly101~\cite{sener2022CVPRassembly101} (bottom). We observe that the model attends the most to frames that are strongly indicative of whether the step will ultimately be executed correctly or not. For example, in the CaptainCook4D case, the model focuses most on the frame where the knife makes a small cut on the bagel, suggesting that the subsequent slice will follow that cut and become too thin, leading to a mistake. In the Assembly101 example, the base aligns well with the chassis in frame 3, indicating that the step is being executed correctly and leading the model to assign the highest attention to that frame.} \label{fig:mistDet_attn}
\end{figure}

In Fig.~\ref{fig:mistDet_attn}, we show qualitative examples that depict our model’s frame-level attention scores at the exit point for CaptainCook4D~\cite{peddi2024NeurIPScaptaincook4d} (top) and Assembly101~\cite{sener2022CVPRassembly101} (bottom). We observe that the model consistently assigns the highest attention weights to frames that contain strong visual cues about whether the step will ultimately be executed correctly. For instance, in the CaptainCook4D example, the model places the most attention on the frame where the knife creates an initial cut on the bagel. This early cut reveals that the slice is likely to follow that trajectory and become excessively thin, providing an early visual signal that the step will end in a mistake. In the Assembly101 example, the base aligns well with the chassis in the third frame, indicating that the step is being executed correctly and leading the model to assign the highest attention to that frame. Such examples highlight the model’s ability to identify informative frames that provide early evidence about the correctness of the ongoing keystep, while also suggesting its capacity to use these cues to anticipate how the remainder of the step is likely to unfold.

Finally, our model can also anticipate that a step will end in a successful execution by leveraging visual cues that indicate the correctness of the remaining portion. For instance, in row 2, the pepper container is already closed after seasoning the eggs, suggesting that the actor is unlikely to add excess pepper. 
By identifying such cues early in the streaming clip, the model can infer the (likely) eventual correctness of the step without needing to observe the entire sequence.

We also observe a common failure mode of our model. It sometimes predicts a step as correct when the observed portion of the clip shows a repetitive action being performed correctly, but the actor begins executing the action incorrectly after the model has already exited. Additionally, the model may incorrectly flag a mistake when a step appears to be performed clumsily but must be executed in that manner due to the nature of the action or the objects involved. In such cases, the model predicts a mistake even though the step is executed correctly. These failures arise when the observed portion of the clip is not representative of the remainder of the step containing non-standard object and action dynamics, and represents well the ``no free lunch" principle of trying to skip seeing parts of the video while still confidently anticipating its results.

See Supp.~video (Sec.~\ref{sec:supp_vid}) for more qualitative examples. 

%% file: sections/conclusion.tex
\section{Conclusion}\label{sec:conclusion}
We introduced the task of early mistake detection in streaming videos, where the goal is to determine the correctness of a keystep in a procedural activity while observing only a minimal fraction of the video. To address this challenge, we proposed a framework that combines a future-anticipating mistake detector with a reinforcement learning policy that adaptively decides when to stop observing and produce a prediction. Experiments on two diverse real-world procedural video datasets show that our approach achieves superior mistake detection accuracy while substantially reducing the amount of video that must be observed compared to prior methods. In future work, we plan to extend our model to handle open-world mistakes, including fine-grained mistake classification.

%% file: sections/supp.tex
\section{Supplementary material}\label{sec:supp}

In this supplementary material we provide additional details about:
\begin{itemize}
    \item Video for qualitatively illustrating  (Sec.~\ref{sec:supp_vid}), as mentioned in `Qualitative examples' 
    in Sec.~\ref{sec:experiments}. 
    \item Analysis of the impact of random seeds on our model performance (Sec.~\ref{sec:supp_analysis_randomSeed})
     \item Additional implementation details (Sec.~\ref{sec:supp_implementation}), as referenced 
     in Sec.~\ref{sec:exp_setup}.
\end{itemize}

\subsection{Supplementary video}~\label{sec:supp_vid}
Tne supplementary video, available at \url{https://vision.cs.utexas.edu/projects/mist_exit}, qualitatively illustrates our task, Early Detection of Mistakes in Procedural Videos, and our MistExit method for tackling this task. We also show successful predictions by our model across both datasets, CaptainCook4D~\cite{peddi2024NeurIPScaptaincook4d} and Assembly101~\cite{sener2022CVPRassembly101}. Finally, we illustrate our model's failure cases 
(Sec.~\ref{sec:results})
with qualitative examples.

\subsection{Impact of random seed on our model performance}~\label{sec:supp_analysis_randomSeed}

\begin{table}[!t]
  \centering
  \setlength{\tabcolsep}{4pt}
    \begin{tabular}{l c c}
    \toprule
     Model & AP \% & OR \% \\
     \midrule
    FastForward~\cite{fan2018IJCAIfastforward} & 56.2 & 5.0\\
    \textbf{Ours} & \textbf{56.8} & \textbf{3.7}\\
    \bottomrule
  \end{tabular}
\caption{Early mistake detection results for our method and the state-of-the-art RL baseline FastForward~\cite{fan2018IJCAIfastforward}, averaged across 3 random seeds. Higher AP, lower OR is better}
  \label{tab:supp_analysis_randomSeed}
\end{table}

In Sec.~\ref{sec:experiments}, 
we evaluated our method using a single random seed. Here, we further evaluate the model with two additional seeds, report the mean performance across the three seeds---one from the main paper and two from this analysis---and compare it against the state-of-the-art RL baseline FastForward~\cite{fan2018IJCAIfastforward}. The results are shown in Table~\ref{tab:supp_analysis_randomSeed}. Our method consistently outperforms the baseline while achieving better efficiency, demonstrating that our design is robust to different random seed initializations.

\subsection{Additional implementation details}~\label{sec:supp_implementation}

Here, we provide our implementation details in addition to what we provided 
in main (Sec.~\ref{sec:exp_setup}). 

For both our mistake detector $\mc{D}$ and policy $\pi$, we use I3D~\cite{carreira2017CVPRi3d} and TSM~\cite{lin2019ICCVtsm} encoders to embed video frames from CC4D and Assembly101, respectively, since these datasets provide such precomputed features and use them to benchmark several tasks introduced alongside the datasets.

For our mistake detector, we project the visual feature inputs~\cite{lin2019ICCVtsm, carreira2017CVPRi3d} into 2048-dimensional features using a single linear layer. Next, we use a one-layer Transformer encoder~\cite{flaborea2024CVPRprego, vaswani2017NIPSattn} 
(Sec.~\ref{sec:detector}) 
with a hidden dimensionality of 1024 to aggregate these features. Finally, to obtain the mistake label estimates and the anticipated features, we apply a single linear layer with the appropriate output dimensionality.

For our policy, we encode its visual inputs and mistake detection score inputs 
(Sec.~\ref{sec:policy}) 
using a 3-layer MLP with ReLU~\cite{nair2010ICMLrelu} activations. The weights of these layers are initialized using Kaiming-normal initialization~\cite{he2015delving}. Our policy network employs a one-layer bidirectional GRU~\cite{cho2014EMNLPgru, chung2014NIPSWgru} with 512 hidden units. The actor and critic networks each consist of a single fully connected layer. To train our policy using PPO~\cite{schulman2017ArXiVppo} 
(Sec.~\ref{sec:exp_setup}), 
we weight the action loss by 1.0, the value loss by 0.5, and the entropy loss by 0.2.

Finally, as mentioned 
in Sec.~\ref{sec:intro},
we will release our code and data to ensure reproducibility.

%% file: main.bib
@String(CVPR  = {IEEE Conf. Comput. Vis. Pattern Recog.})

@String(ICCV  = {Int. Conf. Comput. Vis.})

@String(ECCV  = {Eur. Conf. Comput. Vis.})

@String(ICML  = {Int. Conf. Mach. Learn.})

@String(AAAI  = {AAAI})

@String(CVPR  = {CVPR})

@String(ICCV  = {ICCV})

@String(ECCV  = {ECCV})

@String(ICML  = {ICML})

@inproceedings{wang2023ICCVholoassist,
  title={Holoassist: an egocentric human interaction dataset for interactive ai assistants in the real world},
  author={Wang, Xin and Kwon, Taein and Rad, Mahdi and Pan, Bowen and Chakraborty, Ishani and Andrist, Sean and Bohus, Dan and Feniello, Ashley and Tekin, Bugra and Frujeri, Felipe Vieira and others},
  booktitle={Proceedings of the IEEE/CVF International Conference on Computer Vision},
  pages={20270--20281},
  year={2023}
}

@inproceedings{sener2022CVPRassembly101,
  title={Assembly101: A large-scale multi-view video dataset for understanding procedural activities},
  author={Sener, Fadime and Chatterjee, Dibyadip and Shelepov, Daniel and He, Kun and Singhania, Dipika and Wang, Robert and Yao, Angela},
  booktitle={Proceedings of the IEEE/CVF Conference on Computer Vision and Pattern Recognition},
  pages={21096--21106},
  year={2022}
}

@article{peddi2024NeurIPScaptaincook4d,
  title={CaptainCook4D: A dataset for understanding errors in procedural activities},
  author={Peddi, Rohith and Arya, Shivvrat and Challa, Bharath and Pallapothula, Likhitha and Vyas, Akshay and Gouripeddi, Bhavya and Zhang, Qifan and Wang, Jikai and Komaragiri, Vasundhara and Ragan, Eric and others},
  journal={Advances in Neural Information Processing Systems},
  volume={37},
  pages={135626--135679},
  year={2024}
}

@inproceedings{grauman2024CVPRegoexo4d,
  title={Ego-exo4d: Understanding skilled human activity from first-and third-person perspectives},
  author={Grauman, Kristen and Westbury, Andrew and Torresani, Lorenzo and Kitani, Kris and Malik, Jitendra and Afouras, Triantafyllos and Ashutosh, Kumar and Baiyya, Vijay and Bansal, Siddhant and Boote, Bikram and others},
  booktitle={Proceedings of the IEEE/CVF Conference on Computer Vision and Pattern Recognition},
  pages={19383--19400},
  year={2024}
}

@InProceedings{Ghoddoosian2023CVPRata,
    author    = {Ghoddoosian, Reza and Dwivedi, Isht and Agarwal, Nakul and Dariush, Behzad},
    title     = {Weakly-Supervised Action Segmentation and Unseen Error Detection in Anomalous Instructional Videos},
    booktitle = {Proceedings of the IEEE/CVF International Conference on Computer Vision (ICCV)},
    month     = {October},
    year      = {2023},
    pages     = {10128-10138}
}

@InProceedings{Lee2024CVPRegoper,
  author    = {Lee, Shih-Po and Lu, Zijia and Zhang, Zekun and Hoai, Minh and Elhamifar, Ehsan},
  title     = {Error Detection in Egocentric Procedural Task Videos},
  booktitle = {Proceedings of the IEEE/CVF Conference on Computer Vision and Pattern Recognition (CVPR)},
  month     = {June},
  year      = {2024},
  pages     = {18655-18666}
}

@article{haneji2024egooops,
  title={Egooops: A dataset for mistake action detection from egocentric videos referring to procedural texts},
  author={Haneji, Yuto and Nishimura, Taichi and Kameko, Hirotaka and Shirai, Keisuke and Yoshida, Tomoya and Kajimura, Keiya and Yamamoto, Koki and Cui, Taiyu and Nishimoto, Tomohiro and Mori, Shinsuke},
  journal={arXiv preprint arXiv:2410.05343},
  year={2024}
}

@inproceedings{schoonbeek2024WACVindustreal,
  title={Industreal: A dataset for procedure step recognition handling execution errors in egocentric videos in an industrial-like setting},
  author={Schoonbeek, Tim J and Houben, Tim and Onvlee, Hans and Van der Sommen, Fons and others},
  booktitle={Proceedings of the IEEE/CVF Winter Conference on Applications of Computer Vision},
  pages={4365--4374},
  year={2024}
}

@inproceedings{qian2022CVPRsvip,
  title={Svip: Sequence verification for procedures in videos},
  author={Qian, Yicheng and Luo, Weixin and Lian, Dongze and Tang, Xu and Zhao, Peilin and Gao, Shenghua},
  booktitle={Proceedings of the IEEE/CVF Conference on Computer Vision and Pattern Recognition},
  pages={19890--19902},
  year={2022}
}

@INPROCEEDINGS{jang2019ICCVWepictent,
  author={Jang, Youngkyoon and Sullivan, Brian and Ludwig, Casimir and Gilchrist, Iain D. and Damen, Dima and Mayol-Cuevas, Walterio},
  booktitle={2019 IEEE/CVF International Conference on Computer Vision Workshop (ICCVW)}, 
  title={EPIC-Tent: An Egocentric Video Dataset for Camping Tent Assembly}, 
  year={2019},
  volume={},
  number={},
  pages={4461-4469},
  doi={10.1109/ICCVW.2019.00547}}

@inproceedings{flaborea2024CVPRprego,
  title={PREGO: online mistake detection in PRocedural EGOcentric videos},
  author={Flaborea, Alessandro and Di Melendugno, Guido Maria D'Amely and Plini, Leonardo and Scofano, Luca and De Matteis, Edoardo and Furnari, Antonino and Farinella, Giovanni Maria and Galasso, Fabio},
  booktitle={Proceedings of the IEEE/CVF Conference on Computer Vision and Pattern Recognition},
  pages={18483--18492},
  year={2024}
}

@inproceedings{huang2025CVPRamnar,
  title={Modeling Multiple Normal Action Representations for Error Detection in Procedural Tasks},
  author={Huang, Wei-Jin and Li, Yuan-Ming and Xia, Zhi-Wei and Tang, Yu-Ming and Lin, Kun-Yu and Hu, Jian-Fang and Zheng, Wei-Shi},
  booktitle={Proceedings of the Computer Vision and Pattern Recognition Conference},
  pages={27794--27804},
  year={2025}
}

@article{ding2023every,
  title={Every mistake counts in assembly},
  author={Ding, Guodong and Sener, Fadime and Ma, Shugao and Yao, Angela},
  journal={arXiv preprint arXiv:2307.16453},
  year={2023}
}

@article{seminara2024differentiable,
  title={Differentiable task graph learning: Procedural activity representation and online mistake detection from egocentric videos},
  author={Seminara, Luigi and Farinella, Giovanni Maria and Furnari, Antonino},
  journal={arXiv preprint arXiv:2406.01486},
  year={2024}
}

@inproceedings{mazzamuto2025CVPRgazing,
  title={Gazing into missteps: Leveraging eye-gaze for unsupervised mistake detection in egocentric videos of skilled human activities},
  author={Mazzamuto, Michele and Furnari, Antonino and Sato, Yoichi and Farinella, Giovanni Maria},
  booktitle={Proceedings of the Computer Vision and Pattern Recognition Conference},
  pages={8310--8320},
  year={2025}
}

@article{plini2024ti,
  title={TI-PREGO: Chain of Thought and In-Context Learning for Online Mistake Detection in PRocedural EGOcentric Videos},
  author={Plini, Leonardo and Scofano, Luca and De Matteis, Edoardo and di Melendugno, Guido Maria D'Amely and Flaborea, Alessandro and Sanchietti, Andrea and Farinella, Giovanni Maria and Galasso, Fabio and Furnari, Antonino},
  journal={arXiv preprint arXiv:2411.02570},
  year={2024}
}

@INPROCEEDINGS{bloom2014ICPRcluster,
  author={Bloom, Victoria and Makris, Dimitrios and Argyriou, Vasileios},
  booktitle={2014 22nd International Conference on Pattern Recognition}, 
  title={Clustered Spatio-temporal Manifolds for Online Action Recognition}, 
  year={2014},
  volume={},
  number={},
  pages={3963-3968},
  doi={10.1109/ICPR.2014.679}}

@article{kviatkovsky2014CVIUcovariance,
title = {Online action recognition using covariance of shape and motion},
journal = {Computer Vision and Image Understanding},
volume = {129},
pages = {15-26},
year = {2014},
note = {Special section: Advances in Discrete Geometry for Computer Imagery},
issn = {1077-3142},
doi = {https://doi.org/10.1016/j.cviu.2014.08.001},
url = {https://www.sciencedirect.com/science/article/pii/S1077314214001805},
author = {Igor Kviatkovsky and Ehud Rivlin and Ilan Shimshoni}
}

@inproceedings{suarez2021AAAIonline,
  title={Online action recognition},
  author={Su{\'a}rez-Hern{\'a}ndez, Alejandro and Segovia-Aguas, Javier and Torras, Carme and Alenya, Guillem},
  booktitle={Proceedings of the AAAI Conference on Artificial Intelligence},
  volume={35},
  number={13},
  pages={11981--11989},
  year={2021}
}

@inproceedings{de2016ECCVonline,
  title={Online action detection},
  author={De Geest, Roeland and Gavves, Efstratios and Ghodrati, Amir and Li, Zhenyang and Snoek, Cees and Tuytelaars, Tinne},
  booktitle={European Conference on Computer Vision},
  pages={269--284},
  year={2016},
  organization={Springer}
}

@inproceedings{eun2020CVPRlearning,
  title={Learning to discriminate information for online action detection},
  author={Eun, Hyunjun and Moon, Jinyoung and Park, Jongyoul and Jung, Chanho and Kim, Changick},
  booktitle={Proceedings of the IEEE/CVF conference on computer vision and pattern recognition},
  pages={809--818},
  year={2020}
}

@inproceedings{gao2021CVPRwoad,
  title={Woad: Weakly supervised online action detection in untrimmed videos},
  author={Gao, Mingfei and Zhou, Yingbo and Xu, Ran and Socher, Richard and Xiong, Caiming},
  booktitle={Proceedings of the IEEE/CVF conference on computer vision and pattern recognition},
  pages={1915--1923},
  year={2021}
}

@inproceedings{wang2023ICCVmemory,
  title={Memory-and-anticipation transformer for online action understanding},
  author={Wang, Jiahao and Chen, Guo and Huang, Yifei and Wang, Limin and Lu, Tong},
  booktitle={Proceedings of the IEEE/CVF International Conference on Computer Vision},
  pages={13824--13835},
  year={2023}
}

@inproceedings{wang2021ICCVoadtr,
  title={Oadtr: Online action detection with transformers},
  author={Wang, Xiang and Zhang, Shiwei and Qing, Zhiwu and Shao, Yuanjie and Zuo, Zhengrong and Gao, Changxin and Sang, Nong},
  booktitle={Proceedings of the IEEE/CVF International Conference on Computer Vision},
  pages={7565--7575},
  year={2021}
}

@inproceedings{xu2019ICCVtemporal,
  title={Temporal recurrent networks for online action detection},
  author={Xu, Mingze and Gao, Mingfei and Chen, Yi-Ting and Davis, Larry S and Crandall, David J},
  booktitle={Proceedings of the IEEE/CVF international conference on computer vision},
  pages={5532--5541},
  year={2019}
}

@inproceedings{zhao2022real,
  title={Real-time online video detection with temporal smoothing transformers},
  author={Zhao, Yue and Kr{\"a}henb{\"u}hl, Philipp},
  booktitle={European Conference on Computer Vision},
  pages={485--502},
  year={2022},
  organization={Springer}
}

@InProceedings{an2023ICCVmini,
    author    = {An, Joungbin and Kang, Hyolim and Han, Su Ho and Yang, Ming-Hsuan and Kim, Seon Joo},
    title     = {MiniROAD: Minimal RNN Framework for Online Action Detection},
    booktitle = {Proceedings of the IEEE/CVF International Conference on Computer Vision (ICCV)},
    month     = {October},
    year      = {2023},
    pages     = {8377-8387}
}

@INPROCEEDINGS{soomro2016CVPRonline,
  author={Soomro, Khurram and Idrees, Haroon and Shah, Mubarak},
  booktitle={2016 IEEE Conference on Computer Vision and Pattern Recognition (CVPR)}, 
  title={Predicting the Where and What of Actors and Actions through Online Action Localization}, 
  year={2016},
  volume={},
  number={},
  pages={2648-2657},
  doi={10.1109/CVPR.2016.290}}

@INPROCEEDINGS{kang2021ICCVcontext,
  author={Kang, Hyolim and Kim, Kyungmin and Ko, Yumin and Kim, Seon Joo},
  booktitle={2021 IEEE/CVF International Conference on Computer Vision (ICCV)}, 
  title={CAG-QIL: Context-Aware Actionness Grouping via Q Imitation Learning for Online Temporal Action Localization}, 
  year={2021},
  volume={},
  number={},
  pages={13709-13718},
  doi={10.1109/ICCV48922.2021.01347}}

@InProceedings{kim2022ECCVslide,
author="Kim, Young Hwi
and Kang, Hyolim
and Kim, Seon Joo",
editor="Avidan, Shai
and Brostow, Gabriel
and Ciss{\'e}, Moustapha
and Farinella, Giovanni Maria
and Hassner, Tal",
title="A Sliding Window Scheme for Online Temporal Action Localization",
booktitle="Computer Vision -- ECCV 2022",
year="2022",
publisher="Springer Nature Switzerland",
address="Cham",
pages="653--669",
isbn="978-3-031-19830-4"
}

@InProceedings{Shen2024CVPRprotas,
    author    = {Shen, Yuhan and Elhamifar, Ehsan},
    title     = {Progress-Aware Online Action Segmentation for Egocentric Procedural Task Videos},
    booktitle = {Proceedings of the IEEE/CVF Conference on Computer Vision and Pattern Recognition (CVPR)},
    month     = {June},
    year      = {2024},
    pages     = {18186-18197}
}

@inproceedings{ghoddoosian2022CVPRweakly,
  title={Weakly-supervised online action segmentation in multi-view instructional videos},
  author={Ghoddoosian, Reza and Dwivedi, Isht and Agarwal, Nakul and Choi, Chiho and Dariush, Behzad},
  booktitle={Proceedings of the IEEE/CVF Conference on Computer Vision and Pattern Recognition},
  pages={13780--13790},
  year={2022}
}

@INPROCEEDINGS{donahue2024CVPRself,
  author={Donahue, Gerard and Elhamifar, Ehsan},
  booktitle={2024 IEEE/CVF Conference on Computer Vision and Pattern Recognition (CVPR)}, 
  title={Learning to Predict Activity Progress by Self-Supervised Video Alignment}, 
  year={2024},
  volume={},
  number={},
  pages={18667-18677},
  doi={10.1109/CVPR52733.2024.01766}}

@INPROCEEDINGS{ryoo2011ICCVearly,
  author={Ryoo, M. S.},
  booktitle={2011 International Conference on Computer Vision}, 
  title={Human activity prediction: Early recognition of ongoing activities from streaming videos}, 
  year={2011},
  volume={},
  number={},
  pages={1036-1043},
  doi={10.1109/ICCV.2011.6126349}}

@INPROCEEDINGS{wang2020CVPRactive,
  author={Wang, Boyu and Huang, Lihan and Hoai, Minh},
  booktitle={2020 IEEE/CVF Conference on Computer Vision and Pattern Recognition (CVPR)}, 
  title={Active Vision for Early Recognition of Human Actions}, 
  year={2020},
  volume={},
  number={},
  pages={1078-1088},
  doi={10.1109/CVPR42600.2020.00116}}

@article{chi2024infogcn++,
  title={InfoGCN++: Learning representation by predicting the future for online skeleton-based action recognition},
  author={Chi, Seunggeun and Chi, Hyung-Gun and Huang, Qixing and Ramani, Karthik},
  journal={IEEE Transactions on Pattern Analysis and Machine Intelligence},
  year={2024},
  publisher={IEEE}
}

@INPROCEEDINGS{wang2019CVPRprogressive,
  author={Wang, Xionghui and Hu, Jian-Fang and Lai, Jian-Huang and Zhang, Jianguo and Zheng, Wei-Shi},
  booktitle={2019 IEEE/CVF Conference on Computer Vision and Pattern Recognition (CVPR)}, 
  title={Progressive Teacher-Student Learning for Early Action Prediction}, 
  year={2019},
  volume={},
  number={},
  pages={3551-3560},
  doi={10.1109/CVPR.2019.00367}}

@INPROCEEDINGS{cao2013CVPRpartial,
  author={Cao, Yu and Barrett, Daniel and Barbu, Andrei and Narayanaswamy, Siddharth and Yu, Haonan and Michaux, Aaron and Lin, Yuewei and Dickinson, Sven and Siskind, Jeffrey Mark and Wang, Song},
  booktitle={2013 IEEE Conference on Computer Vision and Pattern Recognition}, 
  title={Recognize Human Activities from Partially Observed Videos}, 
  year={2013},
  volume={},
  number={},
  pages={2658-2665},
  doi={10.1109/CVPR.2013.343}}

@InProceedings{lan2014ECCVhiera,
author="Lan, Tian
and Chen, Tsung-Chuan
and Savarese, Silvio",
editor="Fleet, David
and Pajdla, Tomas
and Schiele, Bernt
and Tuytelaars, Tinne",
title="A Hierarchical Representation for Future Action Prediction",
booktitle="Computer Vision -- ECCV 2014",
year="2014",
publisher="Springer International Publishing",
address="Cham",
pages="689--704",
isbn="978-3-319-10578-9"
}

@InProceedings{li2012ECCVaction,
author="Li, Kang
and Hu, Jie
and Fu, Yun",
editor="Fitzgibbon, Andrew
and Lazebnik, Svetlana
and Perona, Pietro
and Sato, Yoichi
and Schmid, Cordelia",
title="Modeling Complex Temporal Composition of Actionlets for Activity Prediction",
booktitle="Computer Vision -- ECCV 2012",
year="2012",
publisher="Springer Berlin Heidelberg",
address="Berlin, Heidelberg",
pages="286--299",
isbn="978-3-642-33718-5"
}

@ARTICLE{kong2016TPAMImax,
  author={Kong, Yu and Fu, Yun},
  journal={IEEE Transactions on Pattern Analysis and Machine Intelligence}, 
  title={Max-Margin Action Prediction Machine}, 
  year={2016},
  volume={38},
  number={9},
  pages={1844-1858},
  doi={10.1109/TPAMI.2015.2491928}}

@InProceedings{kong2014ECCVdisc,
author="Kong, Yu
and Kit, Dmitry
and Fu, Yun",
editor="Fleet, David
and Pajdla, Tomas
and Schiele, Bernt
and Tuytelaars, Tinne",
title="A Discriminative Model with Multiple Temporal Scales for Action Prediction",
booktitle="Computer Vision -- ECCV 2014",
year="2014",
publisher="Springer International Publishing",
address="Cham",
pages="596--611",
isbn="978-3-319-10602-1"
}

@inproceedings{fernando2021CVPRanti,
  title={Anticipating human actions by correlating past with the future with jaccard similarity measures},
  author={Fernando, Basura and Herath, Samitha},
  booktitle={Proceedings of the IEEE/CVF conference on computer vision and pattern recognition},
  pages={13224--13233},
  year={2021}
}

@ARTICLE{hu2019TPAMIsoft,
  author={Hu, Jian-Fang and Zheng, Wei-Shi and Ma, Lianyang and Wang, Gang and Lai, Jianhuang and Zhang, Jianguo},
  journal={IEEE Transactions on Pattern Analysis and Machine Intelligence}, 
  title={Early Action Prediction by Soft Regression}, 
  year={2019},
  volume={41},
  number={11},
  pages={2568-2583},
  doi={10.1109/TPAMI.2018.2863279}}

@INPROCEEDINGS{wang2019CVPRprogress,
  author={Wang, Xionghui and Hu, Jian-Fang and Lai, Jian-Huang and Zhang, Jianguo and Zheng, Wei-Shi},
  booktitle={2019 IEEE/CVF Conference on Computer Vision and Pattern Recognition (CVPR)}, 
  title={Progressive Teacher-Student Learning for Early Action Prediction}, 
  year={2019},
  volume={},
  number={},
  pages={3551-3560},
  doi={10.1109/CVPR.2019.00367}}

@INPROCEEDINGS{zhaoICCV2019residual,
  author={Zhao, He and Wildes, Rick},
  booktitle={2019 IEEE/CVF International Conference on Computer Vision (ICCV)}, 
  title={Spatiotemporal Feature Residual Propagation for Action Prediction}, 
  year={2019},
  volume={},
  number={},
  pages={7002-7011},
  doi={10.1109/ICCV.2019.00710}}

@inproceedings{cai2018AAAItransfer,
author = {Cai, Yijun and Li, Haoxin and Hu, Jian-Fang and Zheng, Wei-Shi},
title = {Action knowledge transfer for action prediction with partial videos},
year = {2019},
isbn = {978-1-57735-809-1},
publisher = {AAAI Press},
url = {https://doi.org/10.1609/aaai.v33i01.33018118},
doi = {10.1609/aaai.v33i01.33018118},
booktitle = {Proceedings of the Thirty-Third AAAI Conference on Artificial Intelligence and Thirty-First Innovative Applications of Artificial Intelligence Conference and Ninth AAAI Symposium on Educational Advances in Artificial Intelligence},
articleno = {995},
numpages = {8},
location = {Honolulu, Hawaii, USA},
series = {AAAI'19/IAAI'19/EAAI'19}
}

@inproceedings{xu2019ACMMMcgan,
author = {Xu, Wanru and Yu, Jian and Miao, Zhenjiang and Wan, Lili and Ji, Qiang},
title = {Prediction-CGAN: Human Action Prediction with Conditional Generative Adversarial Networks},
year = {2019},
isbn = {9781450368896},
publisher = {Association for Computing Machinery},
address = {New York, NY, USA},
url = {https://doi.org/10.1145/3343031.3351073},
doi = {10.1145/3343031.3351073},
booktitle = {Proceedings of the 27th ACM International Conference on Multimedia},
pages = {611–619},
numpages = {9},
location = {Nice, France},
series = {MM '19}
}

@article{wu2021IJCVSpatialTemporalRR,
  title={Spatial–Temporal Relation Reasoning for Action Prediction in Videos},
  author={Xinxiao Wu and Ruiqi Wang and Jingyi Hou and Hanxi Lin and Jiebo Luo},
  journal={International Journal of Computer Vision},
  year={2021},
  volume={129},
  pages={1484 - 1505},
  url={https://api.semanticscholar.org/CorpusID:233904888}
}

@inproceedings{wu2021AAAIAnticipatingFR,
  title={Anticipating Future Relations via Graph Growing for Action Prediction},
  author={Xinxiao Wu and Jianwei Zhao and Ruiqi Wang},
  booktitle={AAAI Conference on Artificial Intelligence},
  year={2021},
  url={https://api.semanticscholar.org/CorpusID:232416180}
}

@inproceedings{foo2022ECCVera,
  title={Era: Expert retrieval and assembly for early action prediction},
  author={Foo, Lin Geng and Li, Tianjiao and Rahmani, Hossein and Ke, Qiuhong and Liu, Jun},
  booktitle={European Conference on Computer Vision},
  pages={670--688},
  year={2022},
  organization={Springer}
}

@inproceedings{stergiou2023CVPRwisdom,
  title={The wisdom of crowds: Temporal progressive attention for early action prediction},
  author={Stergiou, Alexandros and Damen, Dima},
  booktitle={Proceedings of the IEEE/CVF Conference on Computer Vision and Pattern Recognition},
  pages={14709--14719},
  year={2023}
}

@inproceedings{wang2022CVPRadafocusV2,
  title={Adafocus v2: End-to-end training of spatial dynamic networks for video recognition},
  author={Wang, Yulin and Yue, Yang and Lin, Yuanze and Jiang, Haojun and Lai, Zihang and Kulikov, Victor and Orlov, Nikita and Shi, Humphrey and Huang, Gao},
  booktitle={2022 IEEE/CVF Conference on Computer Vision and Pattern Recognition (CVPR)},
  pages={20030--20040},
  year={2022},
  organization={IEEE}
}

@inproceedings{wang2022ECCVadafocusv3,
  title={Adafocusv3: On unified spatial-temporal dynamic video recognition},
  author={Wang, Yulin and Yue, Yang and Xu, Xinhong and Hassani, Ali and Kulikov, Victor and Orlov, Nikita and Song, Shiji and Shi, Humphrey and Huang, Gao},
  booktitle={European Conference on Computer Vision},
  pages={226--243},
  year={2022},
  organization={Springer}
}

@article{wang2024TPAMIuni,
  title={Uni-adafocus: spatial-temporal dynamic computation for video recognition},
  author={Wang, Yulin and Zhang, Haoji and Yue, Yang and Song, Shiji and Deng, Chao and Feng, Junlan and Huang, Gao},
  journal={IEEE Transactions on Pattern Analysis and Machine Intelligence},
  year={2024},
  publisher={IEEE}
}

@inproceedings{ghodrati2021CVPRframeexit,
  title={Frameexit: Conditional early exiting for efficient video recognition},
  author={Ghodrati, Amir and Bejnordi, Babak Ehteshami and Habibian, Amirhossein},
  booktitle={Proceedings of the IEEE/CVF Conference on Computer Vision and Pattern Recognition},
  pages={15608--15618},
  year={2021}
}

@inproceedings{wu2019CVPRadaframe,
  title={Adaframe: Adaptive frame selection for fast video recognition},
  author={Wu, Zuxuan and Xiong, Caiming and Ma, Chih-Yao and Socher, Richard and Davis, Larry S},
  booktitle={Proceedings of the IEEE/CVF Conference on Computer Vision and Pattern Recognition},
  pages={1278--1287},
  year={2019}
}

@inproceedings{fan2018IJCAIfastforward,
  title     = {Watching a Small Portion could be as Good as Watching All: Towards Efficient Video Classification},
  author    = {Hehe Fan and Zhongwen Xu and Linchao Zhu and Chenggang Yan and Jianjun Ge and Yi Yang},
  booktitle = {Proceedings of the Twenty-Seventh International Joint Conference on
               Artificial Intelligence, {IJCAI-18}},
  publisher = {International Joint Conferences on Artificial Intelligence Organization},
  pages     = {705--711},
  year      = {2018},
  month     = {7},
  doi       = {10.24963/ijcai.2018/98},
  url       = {https://doi.org/10.24963/ijcai.2018/98},
}

@inproceedings{he2015delving,
  title={Delving deep into rectifiers: Surpassing human-level performance on imagenet classification},
  author={He, Kaiming and Zhang, Xiangyu and Ren, Shaoqing and Sun, Jian},
  booktitle={Proceedings of the IEEE international conference on computer vision},
  pages={1026--1034},
  year={2015}
}

@inproceedings{nair2010ICMLrelu,
author = {Nair, Vinod and Hinton, Geoffrey E.},
title = {Rectified linear units improve restricted boltzmann machines},
year = {2010},
isbn = {9781605589077},
publisher = {Omnipress},
address = {Madison, WI, USA},
booktitle = {Proceedings of the 27th International Conference on International Conference on Machine Learning},
pages = {807–814},
numpages = {8},
location = {Haifa, Israel},
series = {ICML'10}
}

@inproceedings{cho2014EMNLPgru,
  title={Learning phrase representations using RNN encoder--decoder for statistical machine translation},
  author={Cho, Kyunghyun and Van Merri{\"e}nboer, Bart and Gul{\c{c}}ehre, {\c{C}}a{\u{g}}lar and Bahdanau, Dzmitry and Bougares, Fethi and Schwenk, Holger and Bengio, Yoshua},
  booktitle={Proceedings of the 2014 conference on empirical methods in natural language processing (EMNLP)},
  pages={1724--1734},
  year={2014}
}

@article{chung2014NIPSWgru,
  title={Empirical evaluation of gated recurrent neural networks on sequence modeling},
  author={Chung, Junyoung and Gulcehre, Caglar and Cho, KyungHyun and Bengio, Yoshua},
  journal={arXiv preprint arXiv:1412.3555},
  year={2014}
}

@inproceedings{zhang2022CVPRaction,
  title={Actionformer: Localizing moments of actions with transformers},
  author={Zhang, Chen-Lin and Wu, Jianxin and Li, Yin},
  booktitle={European Conference on Computer Vision},
  pages={492--510},
  year={2022},
  organization={Springer}
}

@article{vaswani2017NIPSattn,
  title={Attention is all you need},
  author={Vaswani, Ashish and Shazeer, Noam and Parmar, Niki and Uszkoreit, Jakob and Jones, Llion and Gomez, Aidan N and Kaiser, {\L}ukasz and Polosukhin, Illia},
  journal={Advances in neural information processing systems},
  volume={30},
  year={2017}
}

@article{majumder2022NEURIPSfew,
  title={Few-shot audio-visual learning of environment acoustics},
  author={Majumder, Sagnik and Chen, Changan and Al-Halah, Ziad and Grauman, Kristen},
  journal={Advances in Neural Information Processing Systems},
  volume={35},
  pages={2522--2536},
  year={2022}
}

@inproceedings{majumder2025ICCVswitch,
  title={Switch-a-View: View Selection Learned from Unlabeled In-the-wild Videos},
  author={Majumder, Sagnik and Nagarajan, Tushar and Al-Halah, Ziad and Grauman, Kristen},
  booktitle={Proceedings of the IEEE/CVF International Conference on Computer Vision},
  pages={11969--11979},
  year={2025}
}

@inproceedings{carreira2017CVPRi3d,
  title={Quo vadis, action recognition? a new model and the kinetics dataset},
  author={Carreira, Joao and Zisserman, Andrew},
  booktitle={proceedings of the IEEE Conference on Computer Vision and Pattern Recognition},
  pages={6299--6308},
  year={2017}
}

@inproceedings{lin2019ICCVtsm,
  title={Tsm: Temporal shift module for efficient video understanding},
  author={Lin, Ji and Gan, Chuang and Han, Song},
  booktitle={Proceedings of the IEEE/CVF international conference on computer vision},
  pages={7083--7093},
  year={2019}
}

@article{loshchilov2017ArXiVadamw,
  title={Decoupled weight decay regularization},
  author={Loshchilov, Ilya and Hutter, Frank},
  journal={arXiv preprint arXiv:1711.05101},
  year={2017}
}

@article{schulman2017ArXiVppo,
  title={Proximal policy optimization algorithms},
  author={Schulman, John and Wolski, Filip and Dhariwal, Prafulla and Radford, Alec and Klimov, Oleg},
  journal={arXiv preprint arXiv:1707.06347},
  year={2017}
}

@article{kingma2014ArXiVadam,
  title={Adam: A method for stochastic optimization},
  author={Kingma, Diederik P and Ba, Jimmy},
  journal={arXiv preprint arXiv:1412.6980},
  year={2014}
}

@inproceedings{majumder2023chat2map,
  title={Chat2map: Efficient scene mapping from multi-ego conversations},
  author={Majumder, Sagnik and Jiang, Hao and Moulon, Pierre and Henderson, Ethan and Calamia, Paul and Grauman, Kristen and Ithapu, Vamsi Krishna},
  booktitle={Proceedings of the IEEE/CVF Conference on Computer Vision and Pattern Recognition},
  pages={10554--10564},
  year={2023}
}

@INPROCEEDINGS{9720526,
  author={Abrash, Michael},
  booktitle={2021 IEEE International Electron Devices Meeting (IEDM)}, 
  title={Creating the Future: Augmented Reality, the next Human-Machine Interface}, 
  year={2021},
  volume={},
  number={},
  pages={1-11},
  doi={10.1109/IEDM19574.2021.9720526}}

@ARTICLE{9933882,
  author={Yang, Lita and Radway, Robert M. and Chen, Yu-Hsin and Wu, Tony F. and Liu, Huichu and Ansari, Elnaz and Chandra, Vikas and Mitra, Subhasish and Beigné, Edith},
  journal={IEEE Micro}, 
  title={Three-Dimensional Stacked Neural Network Accelerator Architectures for AR/VR Applications}, 
  year={2022},
  volume={42},
  number={6},
  pages={116-124},
  doi={10.1109/MM.2022.3202254}}

@inproceedings{bahlCVPR2023affordances,
  title={Affordances from human videos as a versatile representation for robotics},
  author={Bahl, Shikhar and Mendonca, Russell and Chen, Lili and Jain, Unnat and Pathak, Deepak},
  booktitle={Proceedings of the IEEE/CVF Conference on Computer Vision and Pattern Recognition},
  pages={13778--13790},
  year={2023}
}

@inproceedings{kareer2025ICRAegomimic,
  title={Egomimic: Scaling imitation learning via egocentric video},
  author={Kareer, Simar and Patel, Dhruv and Punamiya, Ryan and Mathur, Pranay and Cheng, Shuo and Wang, Chen and Hoffman, Judy and Xu, Danfei},
  booktitle={2025 IEEE International Conference on Robotics and Automation (ICRA)},
  pages={13226--13233},
  year={2025},
  organization={IEEE}
}

@article{hoque2025egodex,
  title={Egodex: Learning dexterous manipulation from large-scale egocentric video},
  author={Hoque, Ryan and Huang, Peide and Yoon, David J and Sivapurapu, Mouli and Zhang, Jian},
  journal={arXiv preprint arXiv:2505.11709},
  year={2025}
}

@inproceedings{doughty2019pros,
  title={The pros and cons: Rank-aware temporal attention for skill determination in long videos},
  author={Doughty, Hazel and Mayol-Cuevas, Walterio and Damen, Dima},
  booktitle={Proceedings of the IEEE/CVF conference on computer vision and pattern recognition},
  pages={7862--7871},
  year={2019}
}

@inproceedings{huang2024egoexolearn,
  title={Egoexolearn: A dataset for bridging asynchronous ego-and exo-centric view of procedural activities in real world},
  author={Huang, Yifei and Chen, Guo and Xu, Jilan and Zhang, Mingfang and Yang, Lijin and Pei, Baoqi and Zhang, Hongjie and Dong, Lu and Wang, Yali and Wang, Limin and others},
  booktitle={Proceedings of the IEEE/CVF Conference on Computer Vision and Pattern Recognition},
  pages={22072--22086},
  year={2024}
}

@article{panchal2024say,
  title={What to say and when to say it: Live fitness coaching as a testbed for situated interaction},
  author={Panchal, Sunny and Bhattacharyya, Apratim and Berger, Guillaume and Mercier, Antoine and B{\"o}hm, Cornelius and Dietrichkeit, Florian and Pourreza, Reza and Li, Xuanlin and Madan, Pulkit and Lee, Mingu and others},
  journal={Advances in Neural Information Processing Systems},
  volume={37},
  pages={75853--75882},
  year={2024}
}

@inproceedings{huh2025vid2coach,
  title={Vid2Coach: Transforming How-To Videos into Task Assistants},
  author={Huh, Mina and Xue, Zihui and Das, Ujjaini and Ashutosh, Kumar and Grauman, Kristen and Pavel, Amy},
  booktitle={Proceedings of the 38th Annual ACM Symposium on User Interface Software and Technology},
  pages={1--24},
  year={2025}
}

@inproceedings{parmar2019and,
  title={What and how well you performed? a multitask learning approach to action quality assessment},
  author={Parmar, Paritosh and Morris, Brendan Tran},
  booktitle={Proceedings of the IEEE/CVF conference on computer vision and pattern recognition},
  pages={304--313},
  year={2019}
}

@article{feng2025evostruggle,
  title={EvoStruggle: A Dataset Capturing the Evolution of Struggle across Activities and Skill Levels},
  author={Feng, Shijia and Wray, Michael and Mayol-Cuevas, Walterio},
  journal={arXiv preprint arXiv:2510.01362},
  year={2025}
}

@inproceedings{pan2025basket,
  title={Basket: A large-scale video dataset for fine-grained skill estimation},
  author={Pan, Yulu and Zhang, Ce and Bertasius, Gedas},
  booktitle={Proceedings of the Computer Vision and Pattern Recognition Conference},
  pages={28952--28962},
  year={2025}
}

@article{yi2025exact,
  title={Exact: A video-language benchmark for expert action analysis},
  author={Yi, Han and Pan, Yulu and He, Feihong and Liu, Xinyu and Zhang, Benjamin and Oguntola, Oluwatumininu and Bertasius, Gedas},
  journal={arXiv preprint arXiv:2506.06277},
  year={2025}
}
